\documentclass[preprint,12pt]{elsarticle}
\usepackage[utf8]{inputenc} % allow utf-8 input
\usepackage[T1]{fontenc}    % use 8-bit T1 fonts
\usepackage{hyperref}       % hyperlin/ks
\usepackage{url}            % simple URL typesetting
\usepackage{booktabs}       % professional-quality tables
\usepackage{microtype}      % microtypography
\usepackage{lipsum}	    	% Can be removed after putting your text content /
\usepackage{multirow}
\usepackage{graphicx}
\usepackage{doi}
\usepackage[section]{placeins}
\usepackage{mathtools}
\usepackage{amsmath}
\usepackage{amssymb}
\usepackage[ruled,vlined]{algorithm2e}
\usepackage{lineno}
\usepackage[dvipsnames]{xcolor}
\usepackage{enumitem}
\usepackage{tabularx} % Required for adjusting table width
\usepackage{array}
\usepackage{subcaption}

\usepackage{siunitx}

\usepackage{{makecell}}
\setcellgapes{4pt} % Adjust the cell padding

\newcolumntype{Y}{>{\centering\arraybackslash}X}
\newcolumntype{M}[1]{>{\centering\arraybackslash}m{#1}}

\setlist[itemize]{noitemsep, topsep=0pt}
\usepackage{amssymb}
\usepackage[a4paper, total={6in, 8in}, margin=1in]{geometry}

\journal{Elsevier}
\usepackage{color}
\usepackage{soul}

\begin{document}

\begin{frontmatter}

\title{Adaptive Distance-Aware Trunk Deep Operator Learning for Long-Span Roadway Bridges}

\author[inst1]{Bilal Ahmed\corref{cor1}}\ead{ba2702@nyu.edu}
\author[inst1,inst2]{Diab W. Abueidda\corref{cor1}}\ead{da3205@nyu.edu}
\author[inst1,inst3]{Waleed El-Sekelly}
\author[inst1]{Tarek Abdoun}
\author[inst1]{Mostafa E. Mobasher\corref{cor1}}\ead{mostafa.mobasher@nyu.edu}

\affiliation[inst1]{
    organization={Civil and Urban Engineering Department},
    addressline={New York University Abu Dhabi}, 
    country={United Arab Emirates}
}

\affiliation[inst2]{
    organization={National Center for Supercomputing Applications},
    addressline={University of Illinois at Urbana-Champaign}, 
    country={United States of America}
}

\affiliation[inst3]{
    organization={Department of Structural Engineering},
    addressline={Mansoura University}, 
    country={Mansoura, Egypt}
}

\cortext[cor1]{Corresponding author}

\begin{abstract}

Long-span roadway bridges exhibit highly localized structural responses under vehicular loading, making repeated finite element (FE) analysis computationally expensive for applications such as influence surface generation and structural digital twins. Existing scientific machine learning approaches struggle to accurately capture these localized responses because most of the structural domain remains near-zero while only small regions exhibit high-gradient behavior. To address this challenge, this study proposes an adaptive-trunk Deep Operator Network (AD-DeepONet) for localized structural response prediction in large-scale bridge systems. The framework dynamically constructs a load-dependent learning domain using a K-nearest neighbors strategy, allowing the network to focus on structural influence zones. The trunk network is further enhanced using distance-aware features that encode the geometric relationship between the load and structural nodes. A physics-based full-field reconstruction is incorporated through a stiffness-informed Schur complement formulation, enabling predictions at adaptive nodes to be extended to the entire structural domain. To enable scalable training, response data are generated using a reduced-order equivalent shell model that preserves the dominant global behavior while significantly reducing computational cost. The proposed framework is validated on both a benchmark bridge model and the real-world Mussafah Bridge. Results show that the method achieves FEM-level accuracy with relative errors below 5\%, while reducing the total response evaluation time (including full-field reconstruction) by approximately 60$\times$; excluding the post-processing reconstruction step, the AD-DeepONet inference is up to four orders of magnitude faster than FEM. In addition, the framework enables rapid generation of full-field responses, influence lines, and influence surfaces under arbitrary vehicular loading configurations, demonstrating strong potential for large-scale bridge analysis and digital twin applications.

\end{abstract}

\begin{highlights}
\item Adaptive distance-aware trunk DeepONet is proposed for localized structural response prediction.
\item KNN-based influence-domain selection enables learning the localized phenomena.
\item Distance-aware trunk features improve representation of localized behavior.
\item Enables rapid generation of influence lines and surfaces without repeated FE simulations.
\item Demonstrated on the Mussafah Bridge, a complex real-world roadway bridge system.
\end{highlights}


\end{frontmatter}


\section{Introduction}\label{intro}

%\subsection{Overview}\label{Intro_Overview}

Long-span roadway bridges, such as the Mussafah Bridge in Abu Dhabi, represent structurally complex systems characterized by large numbers of degrees of freedom, heterogeneous materials, and multi-component configurations. These structures include interacting subsystems such as girders, slabs, piers, and cross-beams, often composed of reinforced and prestressed concrete. A defining feature of such bridges is the highly localized structural response induced by vehicular wheel loads, where deformation is confined to a small \textit{"influence zone"} while the majority of the domain remains nearly unaffected. This combination of structural scale, system complexity, and strong spatial localization makes accurate and repeated analysis particularly challenging, especially for applications such as real-time analysis for digital twins \cite{chiachio2022structural,ye2019digital,ritto2021digital}.

\subsection{Literature Review and Research Gaps}\label{Intro_Lit}
The standard approach for structural analysis is the finite element method (FEM) \cite{brenner2002mathematical,hughes2012finite}, which provides high-fidelity solutions for complex systems. However, for large-scale bridge models, FEM becomes computationally prohibitive when repeated analyses are required under varying loading conditions. This limitation is critical in practical scenarios involving moving or uncertain vehicular loads, where thousands of simulations may be needed.  Recent advances in machine learning have introduced surrogate models capable of approximating structural response with reduced computational cost. In structural engineering, these approaches have been primarily applied to problems such as damage detection \cite{ahmed2023generalized,ahmed2023unveiling,xu2023computer,xu2022real}, structural deflection prediction \cite{liao2023attention,abdu2023assessment}, and system identification \cite{qiu2024damage,deng2025novel}. However, these methods typically focus on discrete outputs (e.g., sensor-level responses) rather than full-field structural behavior, and are commonly based on standard neural networks that learn direct input–output mappings without capturing the underlying functional relationships of the system. This limitation becomes critical in applications such as influence line and influence surface analysis, which are fundamental tools for evaluating moving-load effects in bridges \cite{deng2025novel,zhan2025impact}. These quantities require repeated evaluation of structural response under varying load positions and are widely used in load rating, weigh-in-motion systems, and design verification \cite{zhou2024research,hekivc2025model}. In practice, they are typically obtained through repeated FEM simulations, which is computationally prohibitive for large-scale bridge models \cite{tao2025research}. Although recent studies have explored ML-based approaches for influence surface estimation \cite{deng2025bridge}, these methods remain limited to inverse formulations or require solving governing equations during training.

Operator learning has emerged as a powerful alternative for modeling parametric PDE problems by learning mappings between function spaces. Chen et al. \cite{chen1995universal} established the theoretical foundation for operator approximation, and Lu et al. \cite{lu2021learning} introduced the DeepONet, which employs a branch–trunk architecture to learn mappings from input functions to solution fields. Several extensions have been proposed, including Bayesian DeepONets \cite{moya2023bayesian}, POD-DeepONets \cite{lu2022comprehensive}, attention-based operators \cite{kissas2022learning}, and physics-informed neural operators \cite{moya2023dae}. Other operator-learning frameworks, such as Fourier Neural Operators (FNOs) \cite{li2023fourier,wen2022u} and Graph Neural Operators (GNOs) \cite{li2020multipole,anandkumar2020neural}, have demonstrated strong performance for continuous-domain problems, although they typically assume structured domains or globally distributed responses. While these advances demonstrate improved capability for complex domains and multiple outputs, they have largely been validated on problems with globally distributed responses. Consequently, their application remains largely confined to continuum, low-dimensional problems, and extending these approaches to large-scale, discrete structural systems with highly localized behavior remains a significant challenge.

Building upon these developments, recent studies have begun to explore operator learning for multiple structural applications. For example, DeepONet-based frameworks have been applied to nonlinear structural response prediction, transient dynamics, and large-scale system modeling \cite{he2023novel,he2024predictions,park2026point}, demonstrating the potential of operator learning as an efficient surrogate for high-fidelity simulations. Applications have also emerged in civil infrastructure systems, such as tunneling-induced settlement prediction and offshore structural response analysis \cite{xu2024multi,cao2024deep}, indicating the growing relevance of these methods for real-world engineering problems. In this context, Ahmed et al.~\cite{ahmed2025physics,ahmed2025physics2} introduced stiffness-informed DeepONet formulations with Schur-complement-based domain reduction, demonstrating that physics-guided operator learning can replicate full-field FEM responses for railway bridges. These approaches were developed for systems subjected to distributed loading, where the structural response is relatively global across the domain. However, extending these methods to long-span roadway bridges presents fundamental challenges. Unlike railway bridges, vehicular loading induces highly localized responses that move spatially with load position. As a result, most of the structural domain contains near-zero or low-variation data, making fixed-domain learning strategies inefficient and leading to poor representation of localized high-gradient regions. In addition, real traffic loading involves vehicles with varying axle configurations (e.g., 4-wheel, 6-wheel, and multi-axle trucks), making it difficult to define a unified learning representation. Directly modeling all possible vehicle configurations is computationally intractable. Finally, generating large, high-fidelity training datasets using full three-dimensional FEM requires thousands of simulations of large-scale systems, creating a significant computational bottleneck that limits scalability \cite{herrmann2024deep,silionis2024deep,zhou2025parameter}. These limitations highlight the need for a learning framework that can efficiently capture localized structural behavior while remaining scalable for large bridge systems.

\subsection{Contributions and Paper Structure}\label{Intro_Contri}

To address these challenges, this study proposes an adaptive distance-aware trunk Deep Operator Network (AD-DeepONet) for efficient full-field structural analysis of long-span roadway bridges, with particular focus on the Mussafah Bridge (Abu Dhabi, UAE) and structures of similar design. The framework introduces a load-dependent learning domain, where a set of nearest neighboring nodes is dynamically selected using a K-nearest neighbors (KNN) strategy. This adaptive Schur domain enables the model to focus on regions of dominant structural response (influence zone), improving the learning of highly localized behavior. The trunk network is further enhanced with distance-aware features that encode the geometric relationship between the load and structural nodes. Physical consistency is enforced through stiffness-based reconstruction, where predictions at adaptive nodes are extended to the full domain using a Schur complement formulation, combining data-driven learning with physics-based equilibrium. To handle variability in vehicular loading, the problem is reformulated at the level of individual wheel loads, allowing arbitrary vehicle configurations to be represented through superposition during inference. To address the data generation bottleneck, a reduced-order modeling strategy based on equivalent shell representations is employed, enabling efficient generation of large training datasets while preserving the essential behavior of the full three-dimensional structure. The proposed framework is developed for a synthetic benchmark bridge and the Mussafah Bridge, demonstrating accurate prediction of localized structural responses with significant computational speedup compared to conventional FEM. In addition, the method enables rapid generation of influence lines and surfaces, providing a practical tool for bridge analysis, design, and digital twin applications.

The remainder of the paper is organized as follows. Section~\ref{sec:method} presents the proposed AD-DeepONet framework, including the problem formulation, adaptive domain construction, distance-aware feature representation, and physics-based full-field reconstruction. Section~\ref{sec:frameworkk} introduces the general computational and evaluation framework, including baseline operator learning methods, scalable data generation, wheel-level load representation, and influence surface generation. Sections~\ref{sec:toy} and \ref{sec:mussafah} present the benchmark bridge and Mussafah Bridge results, respectively. Finally, Section~\ref{sec:conclusion} summarizes the main findings, discusses limitations, and outlines future research directions.

\section{Adaptive Operator Learning for Localized Structural Response}\label{sec:method}

\subsection{Problem Formulation: Localized Operator Learning in Bridge Systems}

\begin{figure}[!htbp]
\centering
\includegraphics[width=1\textwidth]{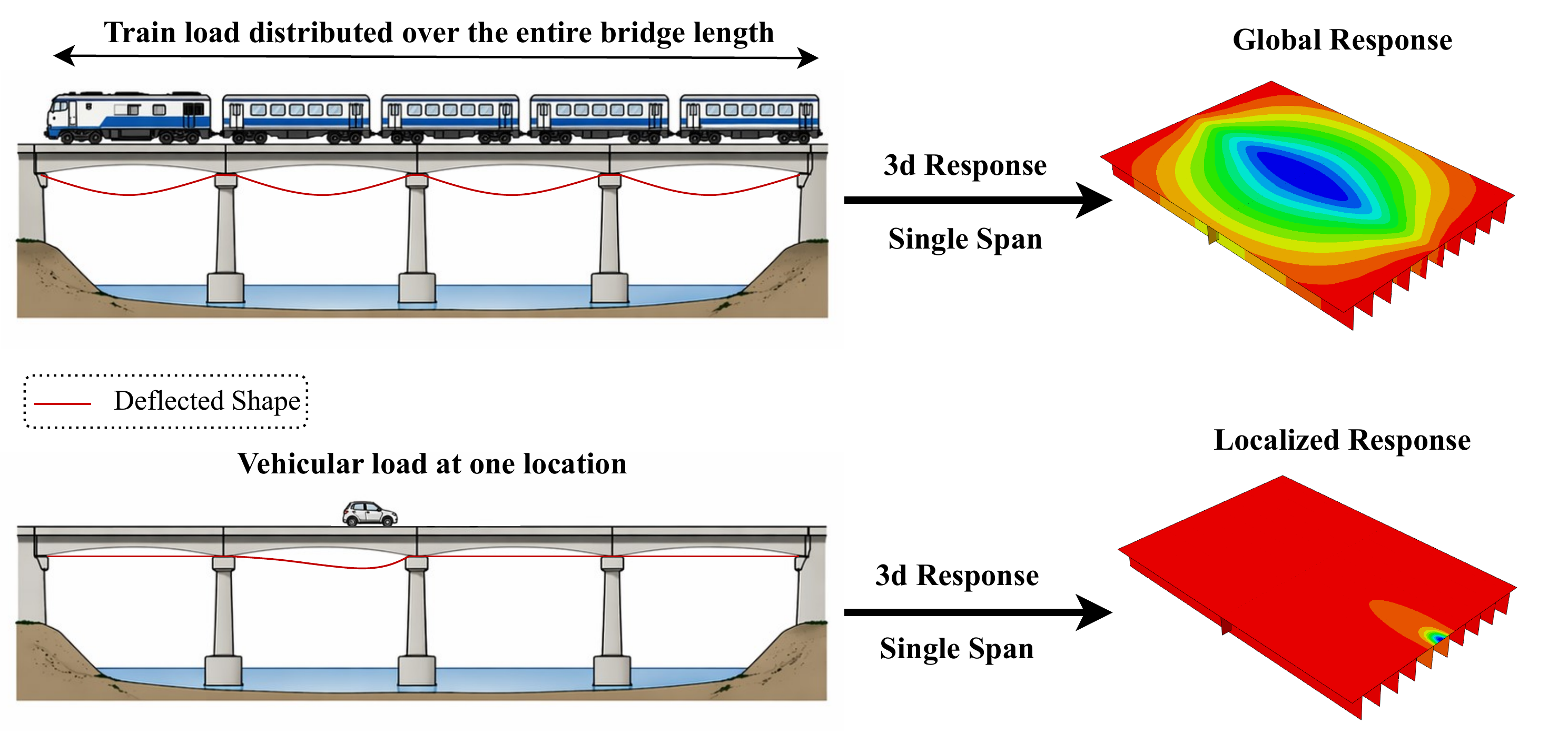}
\caption{Comparison between globally distributed response under railway loading and highly localized response under vehicular loading in a multi-span bridge. Under vehicular loading, deformation is confined to the loaded span, with negligible response in adjacent spans.}
\label{fig:local_global_detailed}
\end{figure}

Long-span roadway bridges, such as the Mussafah Bridge, represent structurally complex systems characterized by large numbers of degrees of freedom, heterogeneous material distributions, and multi-component configurations. Under vehicular loading, these structures exhibit highly localized structural responses, where significant deformation is confined to small regions near the load application point, while the majority of the domain remains nearly unchanged. Figure~\ref{fig:local_global_detailed} illustrates a schematic comparison of structural deformation due to full-span loading versus localized vehicular loading.

The structural behavior is governed by the global equilibrium equation in its matrix form:
\begin{equation}
\mathbf{K}\mathbf{u} = \mathbf{f}
\label{eq:governing_full}
\end{equation}
where $\mathbf{K} \in \mathbb{R}^{n \times n}$ denotes the global stiffness matrix, $\mathbf{u} \in \mathbb{R}^{n}$ represents the displacement and rotation vector, and $\mathbf{f} \in \mathbb{R}^{n}$ corresponds to the applied load vector. The objective of this study is to learn a parametric operator:
\begin{equation}
\mathcal{G} : (F, x_f, z_f) \rightarrow \mathbf{u}(\mathbf{x})
\label{eq:operator_full}
\end{equation}
which maps the load magnitude and spatial location to the full-field structural response. However, for long-span roadway bridges, the target field $\mathbf{u}(\mathbf{x})$ is highly non-uniform and spatially sparse. Specifically, for most $\mathbf{x} \in \Omega$, the response magnitude satisfies:
\begin{equation}
\|\mathbf{u}(\mathbf{x})\| \approx 0, \quad \text{for } \mathbf{x} \notin \mathcal{N}(\mathbf{x}_f)
\end{equation}
where $\mathcal{N}(\mathbf{x}_f)$ denotes a small neighborhood around the load location. This imbalance significantly degrades the ability of data-driven models to capture localized structural behavior.

\subsection{Proposed Adaptive Distance-Aware Trunk DeepONet Framework}

To address these limitations, this study proposes an AD-DeepONet, in which the learning domain dynamically adapts to the load location, ensuring that the network focuses on regions of dominant structural activity.

\subsubsection{Adaptive Schur Domain Construction}\label{sec:adpativesection}

\begin{figure}[!htbp]
\centering
\includegraphics[width=1\textwidth]{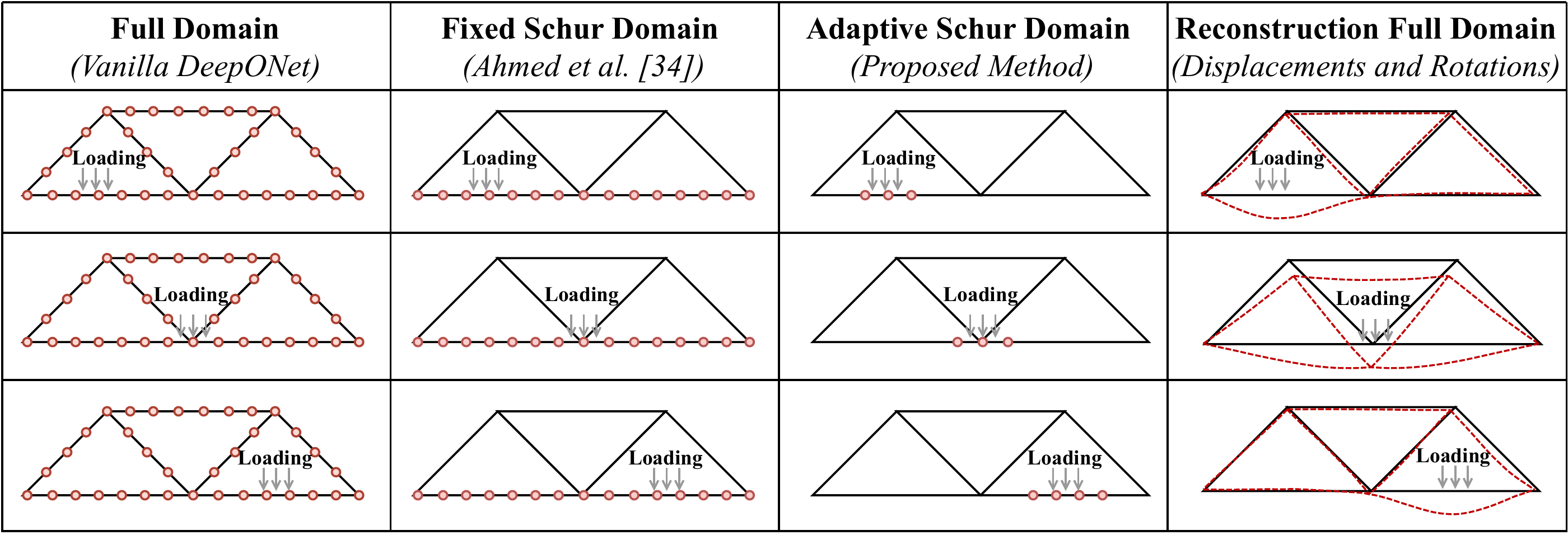}
\caption{Illustration of full-domain, fixed-domain, and adaptive-domain learning strategies under shifting localized loads. Full-domain learning considers the entire structure, fixed-domain learning uses a predefined reduced region, while the adaptive Schur domain dynamically follows the localized influence zone to focus learning on structurally active regions.}
\label{fig:fixedvsadaptive}
\end{figure}

The proposed framework dynamically constructs a load-dependent adaptive domain based on the structural influence zone surrounding the applied load. For each loading scenario, the reduced learning domain is automatically selected according to the load location, enabling the network to focus on regions of dominant structural response. The adaptive domain is defined as:
\begin{equation}\label{Eq:adaptive}
\mathcal{X}_K = \text{KNN}(\mathbf{x}_f)
\end{equation}
where $\mathcal{X}_K$ represents the set of $K$ nearest nodes to the load location $\mathbf{x}_f$. Since localized bridge responses are concentrated near the applied load, this adaptive selection ensures that the selected nodes remain within the active influence zone, while the learning domain dynamically moves with the load position (Figure~\ref{fig:fixedvsadaptive} \textbf{Adaptive Schur Domain}). Furthermore, the model avoids training on irrelevant near-zero regions. This formulation transforms the learning problem from a global operator approximation to a localized operator approximation, significantly improving efficiency and accuracy.

\subsubsection{Distance-Aware Feature Representation}\label{sec:distacneawarefeatures}

To better represent the geometric relationship between the applied load and the structural response region, the trunk input is augmented with distance-aware features:

\begin{equation}
\phi(\mathcal{X}_K) =
\left[
\mathbf{x}_k,\;
\mathbf{x}_k - \mathbf{x}_f,\;
\frac{\mathbf{x}_k - \mathbf{x}_f}{\|\mathbf{x}_k - \mathbf{x}_f\|},\;
\|\mathbf{x}_k - \mathbf{x}_f\|
\right]
\label{eq:Feat}
\end{equation}

Each component in Eq.~\eqref{eq:Feat} provides complementary geometric information regarding the load–structure interaction:

\begin{itemize}
    \item $\mathbf{x}_k$: retains the global spatial position of each node, allowing the network to distinguish different structural regions and preserve the overall bridge geometry.

    \item $\mathbf{x}_k - \mathbf{x}_f$: represents the relative position between the structural node and the load location, effectively shifting the reference frame to a load-centered coordinate system.

    \item $\frac{\mathbf{x}_k - \mathbf{x}_f}{\|\mathbf{x}_k - \mathbf{x}_f\|}$: encodes directional information independent of distance magnitude, enabling the network to distinguish response patterns occurring in different spatial directions relative to the applied load.

    \item $\|\mathbf{x}_k - \mathbf{x}_f\|$: provides a scalar measure of the distance from the load location, which is important for capturing the spatial decay characteristics of localized structural responses.
\end{itemize}

\begin{figure}[!htbp]
\centering
\includegraphics[width=1\textwidth]{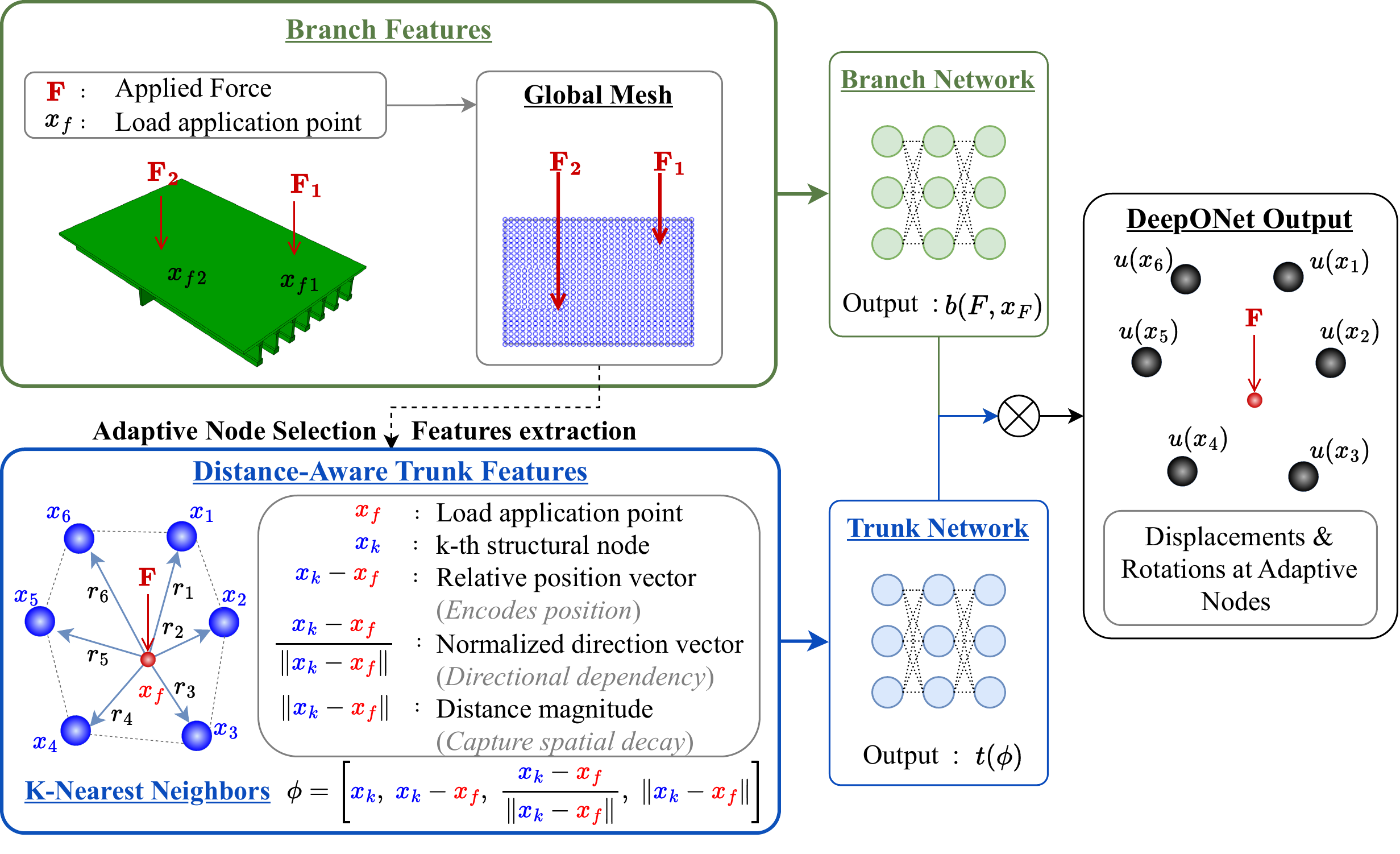}
\caption{Integration of the proposed distance-aware feature representation within the AD-DeepONet framework. The applied wheel load and its location are mapped onto the global mesh, where KNN-based adaptive node selection defines the local Schur domain around the load position. Distance-aware trunk features are constructed for the selected nodes and combined with branch features within the DeepONet architecture to predict localized displacement and rotational responses.}
\label{fig:feature_detailed}
\end{figure}

Together, these features separate spatial information into global position, relative location, direction, and distance. This enriched representation embeds the geometry of load–structure interaction (Figure~\ref{fig:feature_detailed}) and improves the network’s ability to capture spatially varying response patterns. Similar feature augmentation strategies have also been shown to improve DeepONet performance in previous studies \cite{haghighat2024deeponet}.

In addition to the above components, alternative feature augmentations such as inverse distance ($1/\|\mathbf{x}_k - \mathbf{x}_f\|$) and logarithmic distance transformations were also investigated. However, these additional features did not lead to any consistent improvement in predictive accuracy for the present structural system. In some cases, they even introduced numerical instability due to the strong variation near the load location. Therefore, they were not included in the final formulation. It is important to note that the proposed feature construction is general and modular in nature. The trunk representation can be extended with additional problem-specific features when required, making the framework flexible for different structural configurations and loading scenarios.

\subsubsection{Adaptive DeepONet Formulation}

The DeepONet architecture consists of a branch network encoding load parameters and a trunk network processing adaptive features. The output is computed as:
\begin{equation}
u(\mathbf{x}_k) = \sum_{i=1}^{p} b_i(F, x_f, z_f)\; t_i(\phi(\mathcal{X}_K))
\end{equation}

where $b_i$ and $t_i$ denote the branch and trunk outputs, respectively. Unlike conventional DeepONet, the trunk input here is both adaptive and feature-enriched (Figure \ref{fig:feature_detailed}). These modifications enable the network to learn localized operator mappings with significantly improved accuracy.

The model is trained using a data-driven loss function evaluated only over the adaptive Schur domain. For a given training sample $i$, let $\mathcal{X}_K^{(i)}$ denote the selected set of adaptive nodes. The loss is defined as:
\begin{equation}
\mathcal{L}_{\text{Adaptive}} = \frac{1}{N} \sum_{i=1}^{N} \frac{1}{|\mathcal{X}_K^{(i)}|} \sum_{\mathbf{x}_k \in \mathcal{X}_K^{(i)}} 
\left\| u^{\text{FE}}_i(\mathbf{x}_k) - u_i(\mathbf{x}_k) \right\|^2
\label{eq:loss}
\end{equation}

where $u_i(\mathbf{x}_k)$ is the predicted response and $u^{\text{FE}}_i(\mathbf{x}_k)$ is the corresponding finite element solution.

By restricting the loss evaluation to the adaptive domain, the training process focuses on regions of dominant structural response, avoiding the bias introduced by near-zero values over the majority of the domain. Additional implementation details regarding the adaptive trunk representation and network operations are provided in Appendix \ref{app:architecture}.

\subsubsection{Physics-Based Full-Field Reconstruction}\label{sec:fullfield}

The proposed framework follows the physics-based reconstruction strategy introduced in Ahmed et al. \cite{ahmed2025physics}. The AD-DeepONet predicts the structural response only in the adaptive domain $\mathcal{X}_K$ (Eq. \ref{Eq:adaptive}), which defines a subset of nodes selected from the influence zone. The corresponding predicted response is denoted by $\mathbf{U}_I$, representing all degrees of freedom (DOFs) associated with the nodes in $\mathcal{X}_K$. The remaining DOFs, corresponding to nodes outside the adaptive domain, are denoted by $\mathbf{U}_N$.

To recover the full-field solution, the global system is partitioned into predicted ($I$) and non-predicted ($N$) degrees of freedom:
\begin{equation}
\begin{bmatrix}
\mathbf{K}_{II} & \mathbf{K}_{IN} \\
\mathbf{K}_{NI} & \mathbf{K}_{NN}
\end{bmatrix}
\begin{bmatrix}
\mathbf{U}_I \\
\mathbf{U}_N
\end{bmatrix}
=
\begin{bmatrix}
\mathbf{F}_I \\
\mathbf{F}_N
\end{bmatrix}
\end{equation}

The unknown responses at the remaining nodes are then recovered using:
\begin{equation}
\mathbf{U}_N = \mathbf{K}_{NN}^{-1} (\mathbf{F}_N - \mathbf{K}_{NI}\mathbf{U}_I)
\label{eq:schur}
\end{equation}

This hybrid formulation enables full-field recovery by combining data-driven predictions over $\mathcal{X}_K$ with physics-based equilibrium, ensuring consistency with the governing system.

\subsection{Computational Pipeline}

\begin{figure}[!htbp]
\centering
\includegraphics[width=1\textwidth]{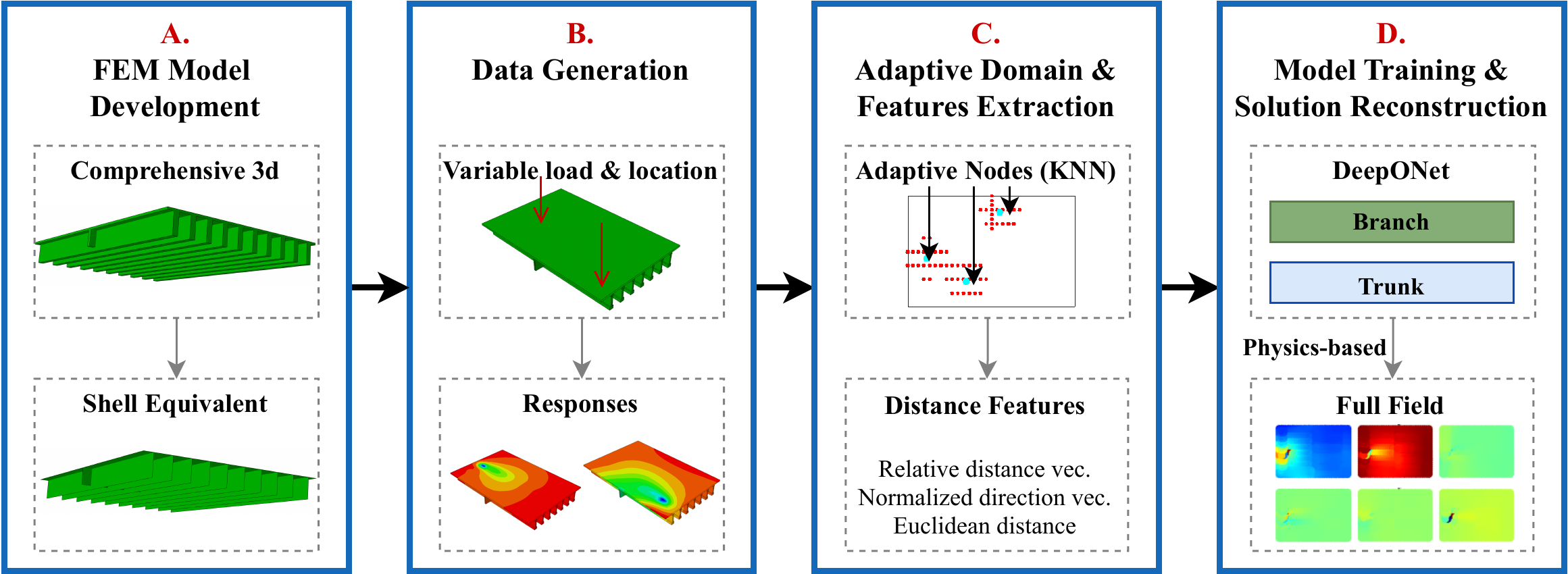}
\caption{Computational pipeline of the proposed AD-DeepONet framework.}
\label{fig:pipeline_detailed}
\end{figure}

The proposed framework integrates model development, data generation, learning, and physics-based reconstruction into a unified workflow (Figure \ref{fig:pipeline_detailed}):

\begin{enumerate}
\item \textbf{Model Development:} A detailed 3D component-level model is constructed, from which an equivalent shell representation is derived for efficient data generation while preserving structural behavior (Figure \ref{fig:pipeline_detailed} \textbf{A}).
\item \textbf{Data Generation:} Structural responses are computed using the equivalent shell-based FEM model under varying wheel load positions(Figure \ref{fig:pipeline_detailed} \textbf{B}).
\item \textbf{Adaptive Domain Selection:} A KNN-based strategy identifies load-dependent nodes, forming the adaptive Schur domain (Figure \ref{fig:pipeline_detailed} \textbf{C}).
\item \textbf{Feature Construction:} Node coordinates are enriched with distance-aware features (Eq.~\ref{eq:Feat}, Figure ~\ref{fig:pipeline_detailed}\textbf{C}).
\item \textbf{Model Training:} DeepONet is trained on the adaptive domain using a data-driven loss (Eq.~\ref{eq:loss},  Figure ~\ref{fig:pipeline_detailed}\textbf{D}).
\item \textbf{Reconstruction:} Full-field responses are recovered via the Schur complement formulation (Eq.~\ref{eq:schur},  Figure ~\ref{fig:pipeline_detailed}\textbf{D}).
\end{enumerate}

The trained model enables fast forward analysis and efficient generation of influence surfaces for arbitrary loading cases.

\section{Computational Framework and Evaluation Methodology}\label{sec:frameworkk}

\subsection{Baseline Operator Learning Methods}

To evaluate the proposed adaptive framework, described in Section \ref{sec:method}, two existing operator learning strategies are considered as baseline approaches: (1) \textit{Method A}: Vanilla DeepONet (Full-Domain Learning) and (2) \textit{Method B}: Fixed-Schur DeepONet (Ahmed et al.~\cite{ahmed2025physics}). Going forward, the proposed method, in Section \ref{sec:method}, will be referred to as \textit{Method C}. 

\subsubsection{Method A: Vanilla DeepONet (Full-Domain Learning)}

In the standard DeepONet formulation, the operator is learned over the entire structural domain:
\begin{equation}
\mathcal{G}: (F,\mathbf{x}_f) \rightarrow \mathbf{u}(\mathbf{x}), \quad \mathbf{x} \in \Omega
\end{equation}
where $(F,\mathbf{x}_f)$ denotes the load magnitude and location, $\Omega$ represents the full structural domain, and $\mathbf{u}(\mathbf{x})$ is the structural response field. The branch network encodes the loading parameters, while the trunk network takes the coordinates of all structural nodes as input and predicts the response at every degree of freedom simultaneously. This formulation directly approximates the global structural response operator without any domain reduction.

Although this approach is general, it becomes inefficient for problems dominated by localized structural behavior. In large-scale bridge systems subjected to concentrated wheel loads, the response is typically confined to a small influence region near the load location:
\begin{equation}
\mathbf{u}(\mathbf{x}) \rightarrow 0, \quad \mathbf{x} \notin \mathcal{X}_{\text{active}}
\end{equation}
where $\mathcal{X}_{\text{active}}$ denotes the localized response region. Consequently, most training samples correspond to near-zero responses, leading to severe data imbalance and reduced learning efficiency. Furthermore, the large output dimensionality substantially increases computational cost during training and inference.

\subsubsection{Method B: Fixed-Schur DeepONet (Ahmed et al.~\cite{ahmed2025physics})}

To reduce the computational complexity associated with full-domain learning, a fixed reduced domain formulation based on the Schur complement was introduced in \cite{ahmed2025physics}. Instead of learning the response over the entire domain $\Omega$, the operator is restricted to a predefined subset of nodes:
\begin{equation}
\mathcal{G}: (F,\mathbf{x}_f) \rightarrow \mathbf{u}(\mathbf{x}), \quad \mathbf{x} \in \mathcal{X}_I
\end{equation}
where $\mathcal{X}_I \subset \Omega$ represents a fixed reduced domain selected prior to training.

The network predicts responses only at the selected Schur nodes, while the remaining degrees of freedom are recovered using a physics-based reconstruction procedure based on the Schur complement formulation. This significantly reduces the output dimensionality and computational cost compared to full-domain learning.

However, the reduced domain remains fixed for all loading scenarios. For localized loading problems, the region of dominant structural response varies with the load position. Consequently, many nodes in the predefined domain may lie outside the active influence zone:
\begin{equation}
\mathbf{u}(\mathbf{x}_k) \rightarrow 0, \quad \mathbf{x}_k \in \mathcal{X}_I
\end{equation}

As a result, the selected domain may not consistently capture the localized response behavior associated with different loading configurations, limiting the robustness and generalization capability of the learned operator.

\subsubsection{Limitations of Existing Operator Learning Approaches}

The above formulations reveal two fundamental limitations in existing operator learning approaches for roadway bridge systems subjected to localized loading. Full-domain learning (Method A) suffers from severe response imbalance and high computational cost due to the large spatial output space. In contrast, fixed-domain reduction (Method B) improves efficiency but lacks adaptability to changing load-dependent response regions.

In practical bridge applications, wheel loads move continuously across the deck, causing the region of structural activity to shift dynamically. Neither global learning nor static domain reduction can efficiently capture this spatially evolving localized behavior. These limitations directly motivate the proposed adaptive-domain formulation, where the learning domain dynamically follows the load location.

\subsection{Scalable Data Generation via Equivalent Shell Modeling}\label{Sec:EqvShell}

A major challenge in applying machine learning to large-scale bridges is the generation of high-fidelity training data. Comprehensive three-dimensional FEM models are computationally prohibitive due to the large number of degrees of freedom and the associated computational cost \cite{hughes2012finite,zienkiewicz2005finite}. To address this, an equivalent shell modeling strategy is adopted. Such approaches are widely used in structural mechanics to reduce computational complexity while preserving the dominant global behavior of large structural systems \cite{reddy2006theory,bathe2006finite}. In this framework, the equivalent shell representation is derived by calibrating a reduced-order model against reference responses obtained from detailed three-dimensional component-level analyses. The calibration is performed to ensure consistency in both static behavior (under unit loading) and dynamic characteristics (natural frequencies). The resulting shell model preserves the essential structural response while significantly reducing the computational complexity. The proposed approach is conceptually aligned with classical substructuring and domain decomposition methods, where large-scale systems are replaced by computationally efficient reduced representations while preserving essential global response characteristics \cite{farhat1991method,farhat2000two}.

\begin{figure}[!htbp]
\centering
\includegraphics[width=1\textwidth]{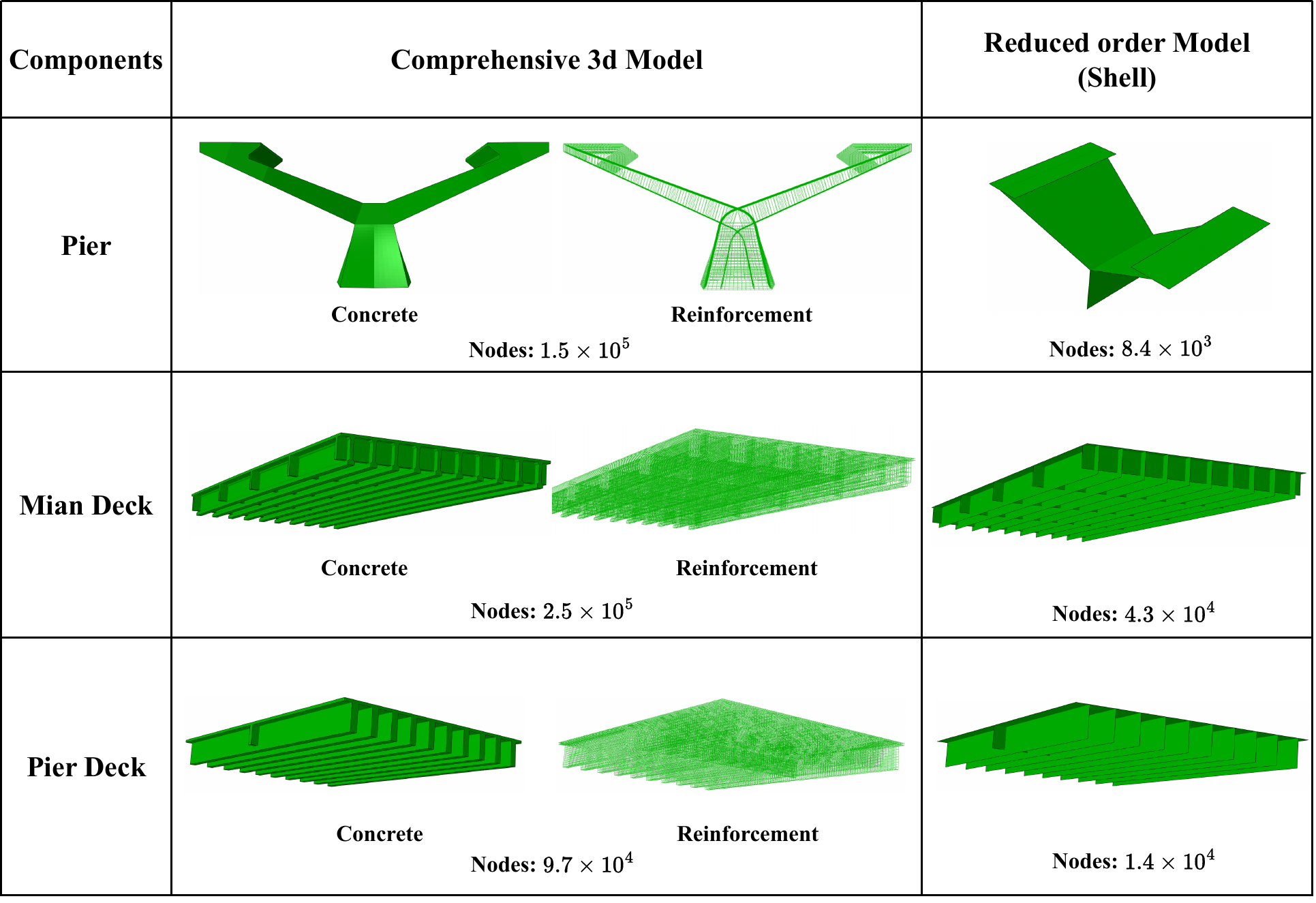}
\caption{Component-level modeling of the Mussafah Bridge, comparing detailed 3D finite element representations with their equivalent shell-based models. The shell idealization significantly reduces the computational cost and model size while preserving the global structural response characteristics.}
\label{fig:Shell}
\end{figure}

This reduced-order modeling strategy significantly decreases the number of nodes while preserving the essential structural response characteristics under loading. As illustrated in Figure~\ref{fig:Shell}, for the Mussafah Bridge component-level modeling, the equivalent shell representation achieves a substantial reduction in computational complexity compared to the detailed 3D model, while maintaining comparable deformation behavior. This reduction enables efficient large-scale dataset generation for training the proposed framework.

\subsection{Load Representation via Wheel-Level Decomposition}\label{sec:wheel_decomposition}

Vehicular loading on roadway bridges involves a wide range of axle layouts and wheel configurations, making direct learning of all possible vehicle arrangements computationally impractical. Standard bridge loading models, such as the AASHTO LRFD design truck and design tandem \cite{aashto2017lrfd}, consist of multiple wheel loads with varying axle spacing and load magnitudes. To develop a unified and scalable representation, the problem is reformulated at the level of individual wheel loads by leveraging structural linearity.

\begin{figure}[!htbp]
\centering
\includegraphics[width=1\textwidth]{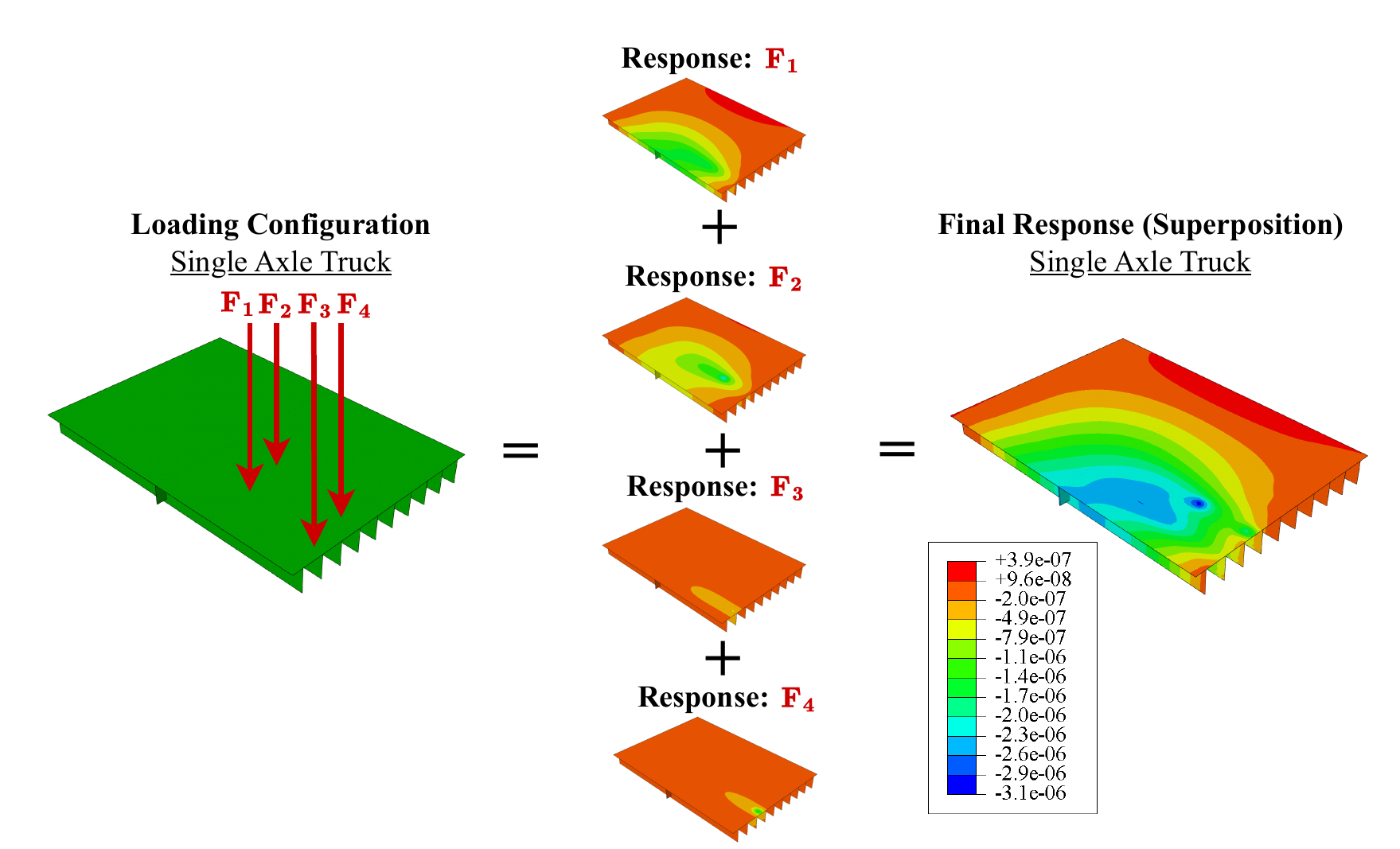}
\caption{Illustration of the superposition-based loading framework, where the design truck load is decomposed into individual wheel loads. The structural response is computed separately for each wheel load and subsequently superimposed to obtain the total bridge response.}
\label{fig:Wheel}
\end{figure}

Instead of directly learning complete vehicle configurations, the operator is trained using single wheel loads applied at different spatial locations. The structural response for a multi-wheel vehicle is then reconstructed through linear superposition of the individual wheel-induced responses. For a vehicle with $n_w$ wheels, the total response is expressed as:
\begin{equation}
\mathbf{u}_{\text{total}}(\mathbf{x}) = \sum_{i=1}^{n_w} \mathcal{G}(F_i, x_i, z_i)(\mathbf{x})
\end{equation}
where $F_i$ denotes the wheel load magnitude, $(x_i,z_i)$ represents the wheel location, and $\mathcal{G}$ is the learned operator corresponding to a single wheel load. For example, a 4-wheel vehicle is decomposed into four individual wheel loads $(F_1,F_2,F_3,F_4)$, whose responses are independently evaluated and superimposed to obtain the total bridge response, as illustrated in Figure~\ref{fig:Wheel}. This formulation enables efficient evaluation of arbitrary vehicle configurations with different axle layouts and load magnitudes without retraining the network. Representative wheel load ranges commonly encountered in bridge traffic loading are summarized in Table~\ref{tab:wheel_loads}.

\begin{table}[!htbp]
\centering
\caption{Representative wheel load ranges for typical vehicular classes}
\label{tab:wheel_loads}
\begin{tabular}{|c|c|}
\hline
\textbf{Vehicle Type} & \textbf{Typical Wheel Load (kN)} \\
\hline
Small Car & 3--5 \\
SUV / Light Pickup & 6--9 \\
Small Truck / Van & 10--15 \\
Medium Duty Truck & 20--30 \\
Heavy Truck & 40--80 \\
Overloaded / Special Vehicle & 80--250 \\
\hline
\end{tabular}
\end{table}

The proposed formulation provides a unified and scalable framework for operator learning under arbitrary vehicular loads. This approach assumes linear structural behavior, which is appropriate for service-level loading conditions in bridge analysis. Under this assumption, the decomposition introduces no loss of accuracy while substantially improving computational efficiency and scalability.

\subsection{Influence Surface Generation}

Influence lines and surfaces quantify the variation of structural response at a fixed observation location due to a moving unit load. Conventionally, their construction requires repeated FEM simulations over many load positions, which becomes computationally expensive for large bridge systems.

Using the trained operator, for a fixed observation location $\mathbf{x}_s$, the influence response due to a unit load at position $\mathbf{x}_f=(x_f,z_f)$ is defined as:
\begin{equation}
I(\mathbf{x}_s,\mathbf{x}_f)=\mathcal{G}(1,x_f,z_f)(\mathbf{x}_s)
\end{equation}
where $\mathcal{G}(1,x_f,z_f)$ denotes the predicted structural response field corresponding to the moving unit load. By systematically varying the load position across the bridge deck, the full influence surface is obtained as:
\begin{equation}
\mathcal{I}(\mathbf{x}_s)=
\left\{
I(\mathbf{x}_s,\mathbf{x}_f)
\;\middle|\;
\mathbf{x}_f \in \Omega
\right\}
\end{equation}

The key advantage of this formulation is that each evaluation of $\mathcal{G}$ is computationally inexpensive compared to FEM, enabling quick generation of influence surfaces. Unlike conventional approaches \cite{deng2025bridge,zhan2025impact}, this method avoids repeated solution of the governing equations and instead relies on the learned operator. Furthermore, complex vehicle configurations can be evaluated using the wheel-level decomposition framework (Section~\ref{sec:wheel_decomposition}), where the total influence response is obtained through linear superposition of individual wheel contributions. This enables rapid generation of influence responses for multi-axle vehicles such as design truck and tandem loading configurations.

\section{Benchmark Bridge (Method Verification)}\label{sec:toy}

To evaluate the proposed AD-DeepONet framework for localized structural response, a large-scale synthetic bridge model is considered. This benchmark is designed to present the key challenges observed in real roadway bridges such as the Mussafah Bridge, including multi-span configuration, structural discontinuities, and highly localized responses under vehicular loading. While the Mussafah Bridge serves as the primary motivation of this study, the synthetic model enables controlled data generation, reproducibility, and systematic evaluation of different learning strategies.

\subsection{Benchmark Structure Description}\label{sec:toy_geometry}

\begin{figure}[!htbp]
    \centering
    \includegraphics[width=1.0\textwidth]{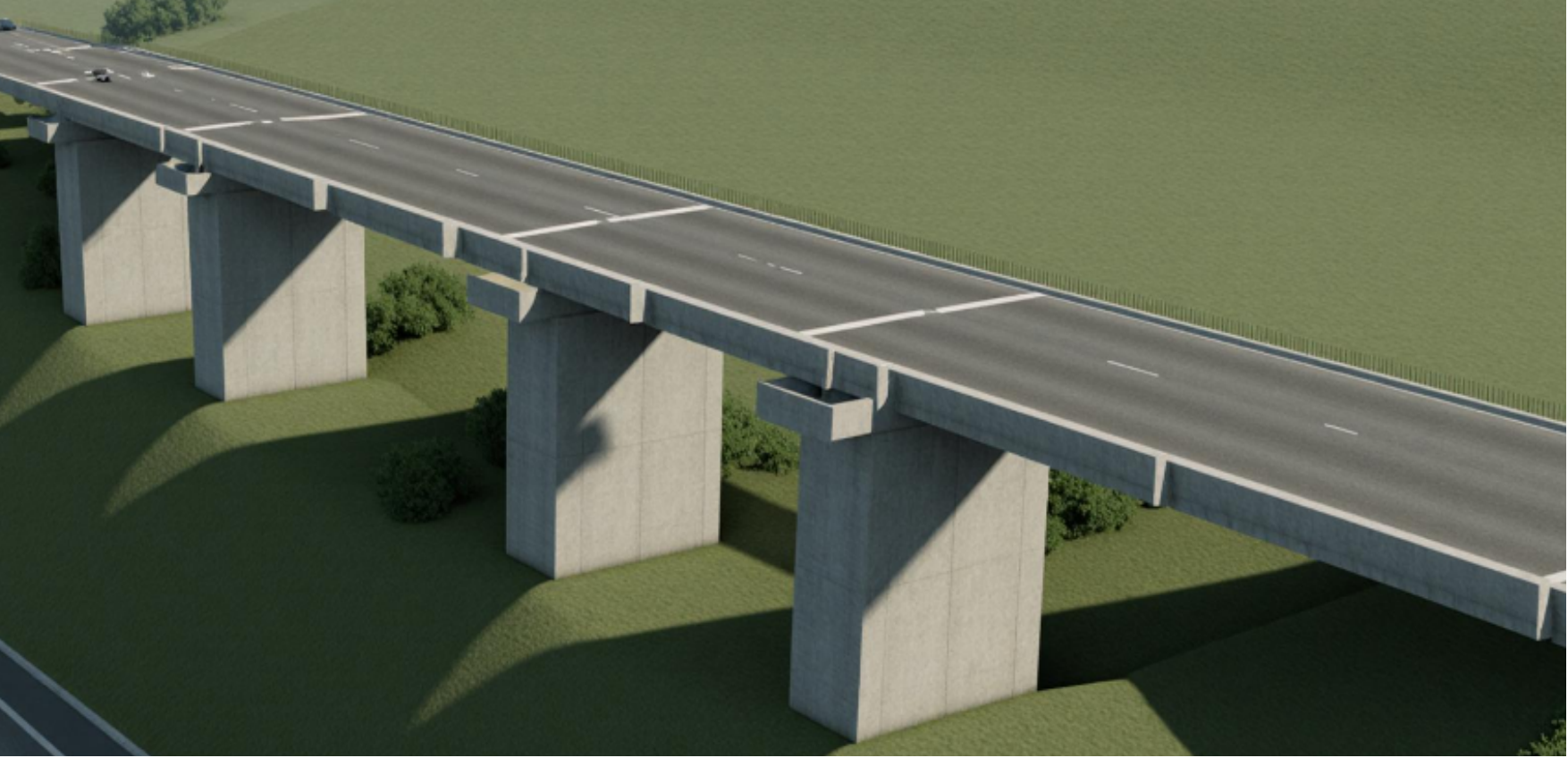}
    \caption{Synthetic benchmark bridge geometry, representing a multi-span roadway bridge with structural discontinuities and unequal span lengths.}
    \label{Benchmark_Geo}
\end{figure}

The benchmark bridge is a multi-span concrete structure with a total length of 105 m and a width of 10 m, consisting of five spans arranged in an alternating configuration (25 m–15 m–25 m–15 m–25 m). This non-uniform layout introduces stiffness discontinuities and span-level segmentation, which are critical for assessing localized response behavior. The superstructure comprises seven longitudinal I-girders spaced at 1.5 m, connected through transverse cross-beams and supporting a 0.16 m thick deck slab. The structure includes expansion joints between spans, resulting in discontinuous load transfer and confinement of structural response within individual spans. The substructure consists of wall-type piers with fixed bases and pin-type connections at the girder–pier interface, while abutments provide end supports. All components are assumed to be constructed from concrete with an elastic modulus of 37,500 MPa, Poisson’s ratio of 0.2, and density of 24 kN/m\(^3\). A conceptual rendering of the bridge geometry is shown in Figure \ref{Benchmark_Geo}, highlighting the multi-span configuration and structural layout.

\subsection{Finite Element Model and Data Generation}\label{sec:fem_data}

A finite element model of the bridge is developed in Abaqus using four-node shell elements (S4R), providing an efficient representation while capturing bending and rotational behavior. All structural components, including girders, deck slab, cross-beams, and piers are modeled using shell formulations. The assembled model consists of 14,157 nodes and 13,784 elements, resulting in 84,942 degrees of freedom. This scale is substantially larger than the KW51 bridge model investigated in our previous study \cite{ahmed2025physics}, which contained approximately 11,000 degrees of freedom. The increase in system size highlights the computational challenges associated with applying operator learning to real bridge systems such as the Mussafah Bridge.

Vehicular loading is represented at the level of individual wheel loads, consistent with the wheel-level decomposition framework described in Section~\ref{sec:wheel_decomposition}. In the FE implementation, each wheel load is applied as an equivalent uniform pressure over a single shell element rather than as a direct nodal force, which improves numerical stability in shell-based discretizations. The adopted mesh size is approximately $475 \,\text{mm} \times 475 \,\text{mm}$, defining the effective load application region within the discretized model. Load magnitudes vary between 1 kN and 250 kN, and load locations are randomly distributed across the bridge deck. Due to the multi-span configuration and structural discontinuities, the resulting structural response remains highly localized within individual spans.

A total of 16,000 static simulations are performed, generating a dataset structured for operator learning:
\begin{itemize}
    \item Input: load magnitude and location $(16000 \times 3)$
    \item Domain: nodal coordinates $(14157 \times 3)$
    \item Output: structural response $(16000 \times 14157 \times 6)$
\end{itemize}

\begin{figure}[!htbp]
    \centering
    \includegraphics[width=0.7\textwidth]{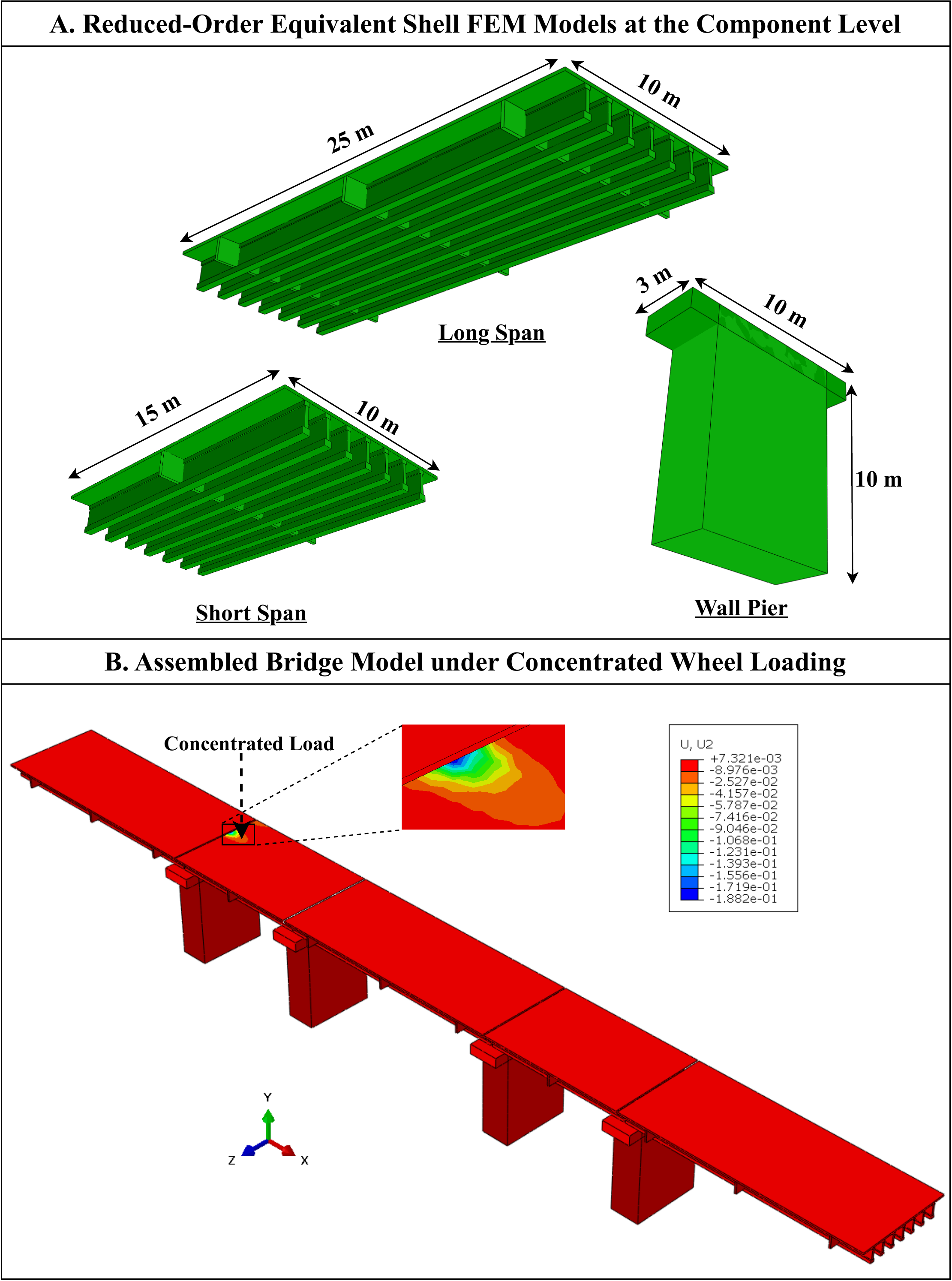}
    \caption{Reduced-order finite element representation adopted for scalable structural analysis. \textbf{A.} Component-level shell models with thickness visualization enabled in Abaqus for enhanced geometric interpretation; the actual computational model employs equivalent shell elements to reduce computational complexity. \textbf{B.} Assembled bridge shell model subjected to a concentrated wheel load.}
    \label{Benchmark_Simulation}
\end{figure}

The six output channels correspond to translational and rotational degrees of freedom $(U_x, U_y, U_z, R_x, R_y, R_z)$. The complete data generation process took more than 10 days, underscoring the computational burden of large-scale bridge simulations. Figure \ref{Benchmark_Simulation} illustrates the finite element model and a representative concentrated-loading case, highlighting the highly localized nature of the structural response.

\subsection{Statistical Characterization of Structural Response}

\begin{figure}[!htbp]
    \centering
    \includegraphics[width=1\textwidth]{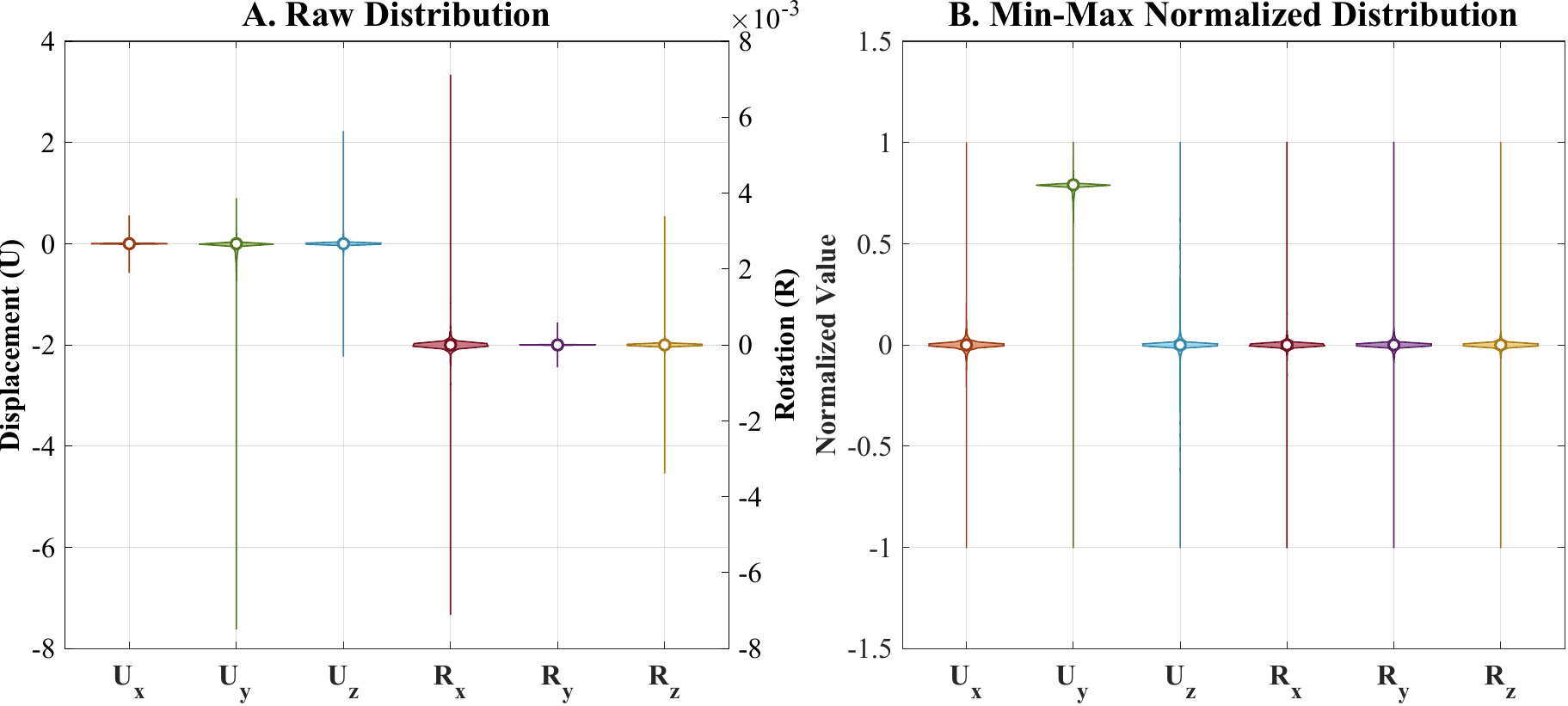}
    \caption{Statistical distribution of the response variables. \textbf{A.} Raw data distributions of the displacement and rotational response components ($U_x$, $U_y$, $U_z$, $R_x$, $R_y$, and $R_z$). \textbf{B.} Corresponding min-max normalized distributions mapped to the range [-1,1], which is commonly adopted in machine learning frameworks for stable training and improved convergence.}
    \label{Benchmark_Distribution}
\end{figure}

To quantify the nature of the dataset, a statistical summary of the response variables is provided in Table~\ref{tab:Statistical}, with distributions illustrated in Figure~\ref{Benchmark_Distribution}. The results clearly indicate a highly localized response behavior. For all variables, the mean and median values are close to zero, while the standard deviation remains small relative to the overall range. This confirms that the majority of nodes experience negligible response, with significant variations confined to a small region near the load application.

This effect is particularly pronounced for rotational degrees of freedom ($R_x$, $R_y$, $R_z$), where responses are nearly zero across most of the domain. Even for displacement components ($U_x$, $U_y$, $U_z$), large extrema coexist with low variance, indicating sparse high-response regions embedded within a predominantly inactive domain. Importantly, this imbalance persists even after standard min–max normalization (Figure \ref{Benchmark_Distribution} \textbf{B}). Such data characteristics pose a fundamental challenge for machine learning models trained over the full domain, as the loss becomes dominated by near-zero regions, leading to poor representation of localized high-gradient responses. These observations highlight the need for a learning strategy that focuses on informative regions of the domain.

\begin{table}[]
\caption{Statistical summary of displacement (mm) and rotational (rad) response variables, including minimum, maximum, mean, median, standard deviation, and range.}
\label{tab:Statistical}
\begin{tabular}{|c|c|c|c|c|c|c|}
\hline
\textbf{Variables} & \textbf{Min} & \textbf{Max} & \textbf{Mean} & \textbf{Median} & \textbf{Std} & \textbf{Range} \\ \hline
\textbf{$U_x$} (mm)       & -5.612e-1    & 5.612e-01    & 2.012e-05     & -4.395e-06      & 9.654e-03    & 1.122          \\
\textbf{$U_y$} (mm)       & -7.605       & 8.877e-01    & -1.038e-02    & 2.529e-04       & 7.669e-02    & 8.493          \\
\textbf{$U_z$} (mm)      & -2.214       & 2.214        & 1.168e-06      & 6.778e-10       & 2.098e-02     & 4.427          \\
\textbf{$R_x$} (rad)       & -7.097e-03   & 7.097e-03    & -1.270e-09    & -4.123e-13      & 2.891e-05    & 1.419e-02      \\
\textbf{$R_y$} (rad)        & -5.715e-04   & 5.715e-04    & 1.114e-09     & 7.142e-13       & 4.313e-06    & 1.143e-03      \\
\textbf{$R_z$} (rad)        & -3.371e-03   & 3.371e-03    & -8.509e-10    & -3.455e-31      & 1.310e-05    & 6.742e-03      \\ \hline
\end{tabular}
\end{table}

\subsection{Method Comparison and Results Discussion} \label{sec:toy_comparison}
\begin{figure}[!htbp]
    \centering
    \includegraphics[width=1\textwidth]{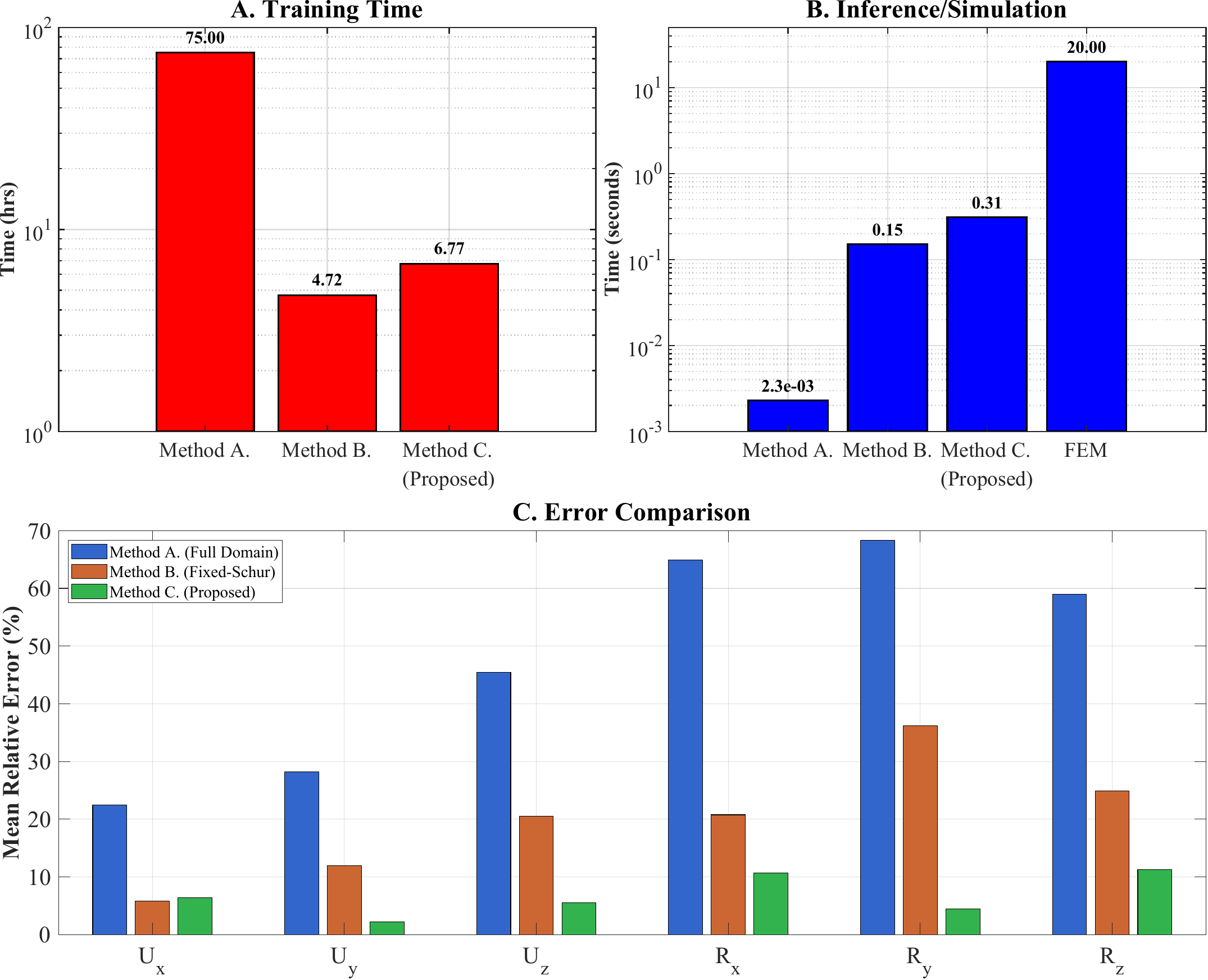}
    \caption{Comprehensive performance comparison of Methods A, B, and C for the benchmark bridge problem. \textbf{A.} Training time comparison. \textbf{B.} Inference time comparison. \textbf{C.} Mean relative error for each structural response component across all evaluated methods.}
\label{Benchmark_Comparison}
\end{figure}

The performance of the three operator learning strategies, namely Method A (Vanilla DeepONet), Method B (Fixed-Schur DeepONet \cite{ahmed2025physics}), and Method C (Proposed AD-DeepONet), is evaluated in terms of prediction accuracy and computational efficiency. A summary of the training cost, inference time, and mean relative errors for all response components is presented in Figure~\ref{Benchmark_Comparison}. The results clearly demonstrate that Method C provides the best balance between accuracy and computational efficiency for localized structural response prediction.

\begin{figure}[!htbp]
    \centering
    \includegraphics[width=1\textwidth]{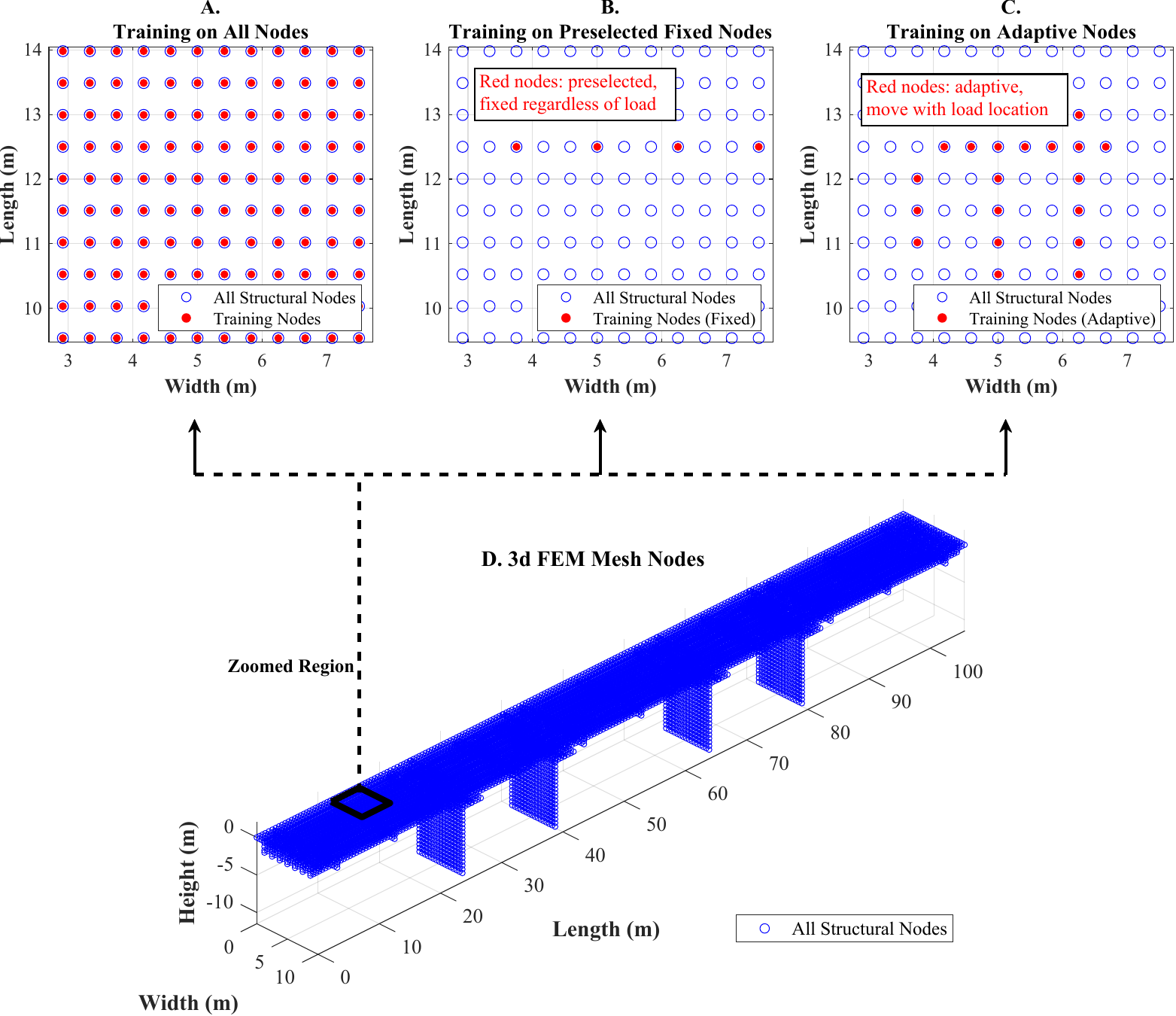}
    \caption{Comparison of training domains adopted for localized response learning. \textbf{A.} Method A (Vanilla DeepONet), where the entire structural domain is used for training and prediction. \textbf{B.} Method B (Fixed-Schur DeepONet \cite{ahmed2025physics}), which employs a predefined fixed Schur domain independent of the load location. \textbf{C.} Proposed Method C (Adaptive-Schur Domain), where the training domain dynamically adapts according to the load position, enabling the selected nodes to spatially align with the moving load. \textbf{D.} Three-dimensional finite element mesh representation of the complete bridge structure, highlighting the full computational domain.}
    \label{Benchmark_AllDomain}
\end{figure}

\paragraph{Method A: Vanilla DeepONet (Full-Domain Learning)}
Method A corresponds to a conventional DeepONet trained over the full spatial domain, where all structural nodes are included in the learning process (Figure~\ref{Benchmark_AllDomain} \textbf{A}). Although this formulation provides a complete representation of the structural domain, its performance degrades significantly for problems dominated by localized responses. As shown in Figure~\ref{Benchmark_All_Error} \textbf{A}, the model exhibits high prediction errors across both displacement and rotational components, with mean errors ranging approximately from 22\% to 68\%. In addition, the large output dimensionality leads to substantially higher training and inference costs (Figure~\ref{Benchmark_Comparison} \textbf{A}) without corresponding improvements in predictive accuracy.

\paragraph{Method B: Fixed-Schur DeepONet (Ahmed et al.~\cite{ahmed2025physics})}
Method B adopts the fixed Schur-domain formulation proposed in \cite{ahmed2025physics}, where a predefined set of 35 nodes, primarily located near the mid-span region, is selected as the reduced learning domain (Figure~\ref{Benchmark_AllDomain} \textbf{B}). Restricting the prediction space significantly reduces the computational cost compared to full-domain learning. However, the predictive performance remains spatially inconsistent. As shown in Figure~\ref{Benchmark_All_Error} \textbf{B}, moderate accuracy is achieved for certain displacement components (e.g., $U_x = 5.8\%$, $U_y = 11.9\%$), while larger errors persist for other components, particularly rotational responses.

\begin{figure}[!htbp]
    \centering
    \includegraphics[width=0.95\textwidth]{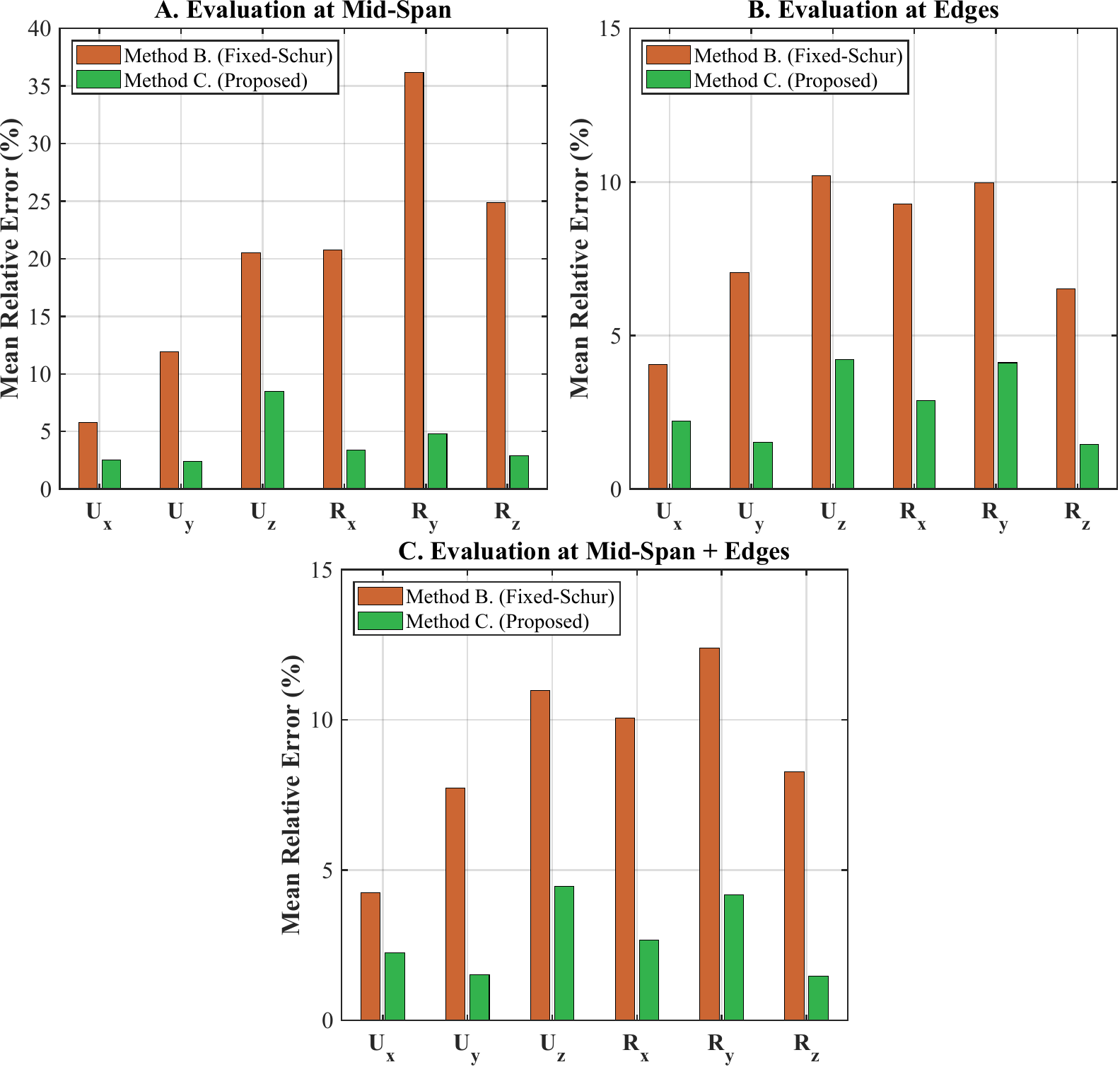}
     \caption{Comparison of full-field reconstruction errors evaluated at different spatial regions for Method B (Fixed-Schur DeepONet) and Method C (Adaptive-Schur DeepONet). \textbf{A.} Error comparison at the bridge mid-span region. \textbf{B.} Error comparison at the edge regions of the bridge deck. \textbf{C.} Error comparison considering the combined mid-span and edge regions.}
    \label{Benchmark_DiffLocation_Error}
\end{figure}

Although the full-field response can be reconstructed using the stiffness-based formulation, the apparent reduction in global error can be misleading (Figure~\ref{Benchmark_All_Error} \textbf{C}). Figure~\ref{Benchmark_DiffLocation_Error} shows that the reconstruction error varies considerably across different spatial regions. Since the reduced domain remains fixed for all loading scenarios, it does not consistently align with the localized influence zone associated with changing load positions, resulting in spatially varying prediction quality.

\paragraph{Method C: AD-DeepONet (Proposed)}
Method C introduces the proposed adaptive Schur-domain formulation, where the learning domain is dynamically constructed for each loading scenario using a KNN-based selection strategy. In this study, 20 neighboring nodes are selected for each sample, resulting in a reduced output space of size $(16000 \times 20 \times 6)$. The neighborhood size is selected based on a parametric study balancing prediction accuracy and computational cost (Appendix~\ref{sec:toy_parametric}). In addition, the trunk input is enriched using the distance-aware feature representation described in Section~\ref{sec:distacneawarefeatures}. Figure~\ref{Benchmark_AllDomain} \textbf{C} illustrates how the adaptive domain continuously aligns with the load location. As shown in Figure~\ref{Benchmark_All_Error} \textbf{D}, the proposed method achieves a substantial reduction in prediction error. Most samples exhibit mean errors below 11\%, with displacement components generally below 6\%. Following stiffness-based full-field reconstruction (Figure~\ref{Benchmark_All_Error} \textbf{E}), the global response errors are further reduced, with most displacement components below 4\%. Slightly larger relative errors observed for $R_y$ are mainly associated with its extremely small response magnitude, where minor absolute deviations produce relatively large percentage errors.

Compared to Method B, the proposed adaptive formulation also demonstrates significantly improved spatial robustness. Figure~\ref{Benchmark_DiffLocation_Error} shows that Method C maintains consistent accuracy across mid-span, edge, and combined regions, whereas the fixed-domain approach exhibits strong spatial variability. These results demonstrate that aligning the learning domain with the localized structural response region is critical for accurate operator learning in bridge systems.

\begin{figure}[!htbp]
    \centering
    \includegraphics[width=0.95\textwidth]{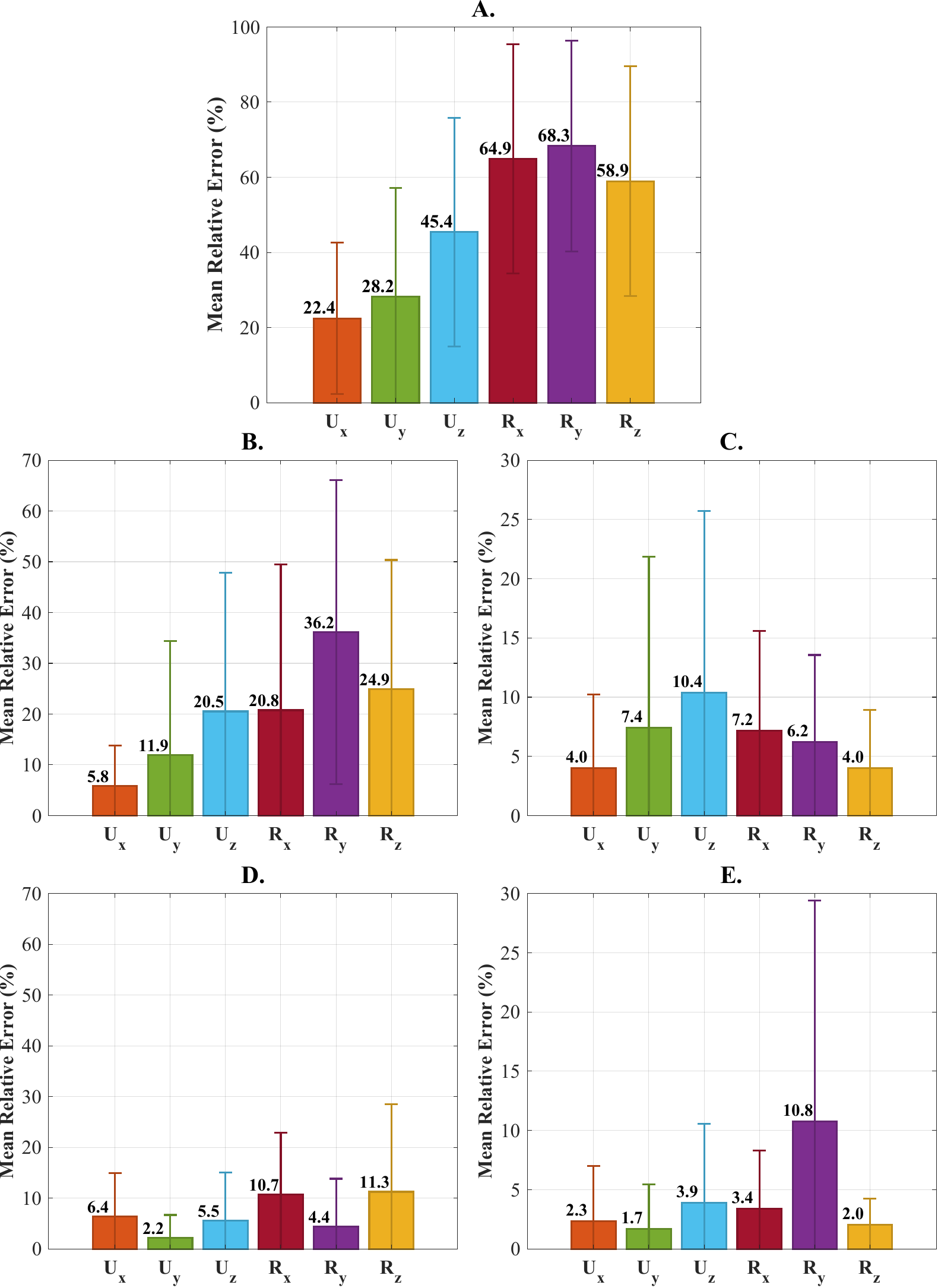}
    \caption{Prediction error comparison for the benchmark bridge problem. \textbf{A.} Full-domain prediction error for Method A (Vanilla DeepONet). \textbf{B.} ML prediction error evaluated at the fixed Schur nodes for Method B (Fixed-Schur DeepONet \cite{ahmed2025physics}). \textbf{C.} Corresponding full-field reconstruction error for Method B obtained using the stiffness-based reconstruction framework described in Section~\ref{sec:fullfield}. \textbf{D.} ML prediction error evaluated at the adaptive Schur nodes for the proposed Method C. \textbf{E.} Corresponding full-field reconstruction error for Method C using the stiffness-based reconstruction framework.}
\label{Benchmark_All_Error}
    \label{Benchmark_All_Error}
\end{figure}

\subsection{Application}
Once trained, the proposed AD-DeepONet enables two categories of structural analysis: (i) rapid forward prediction for a given loading configuration, and (ii) efficient generation of influence lines and influence surfaces through repeated operator evaluations.

\begin{figure}[!htbp]
    \centering
    \includegraphics[width=0.85\textwidth]{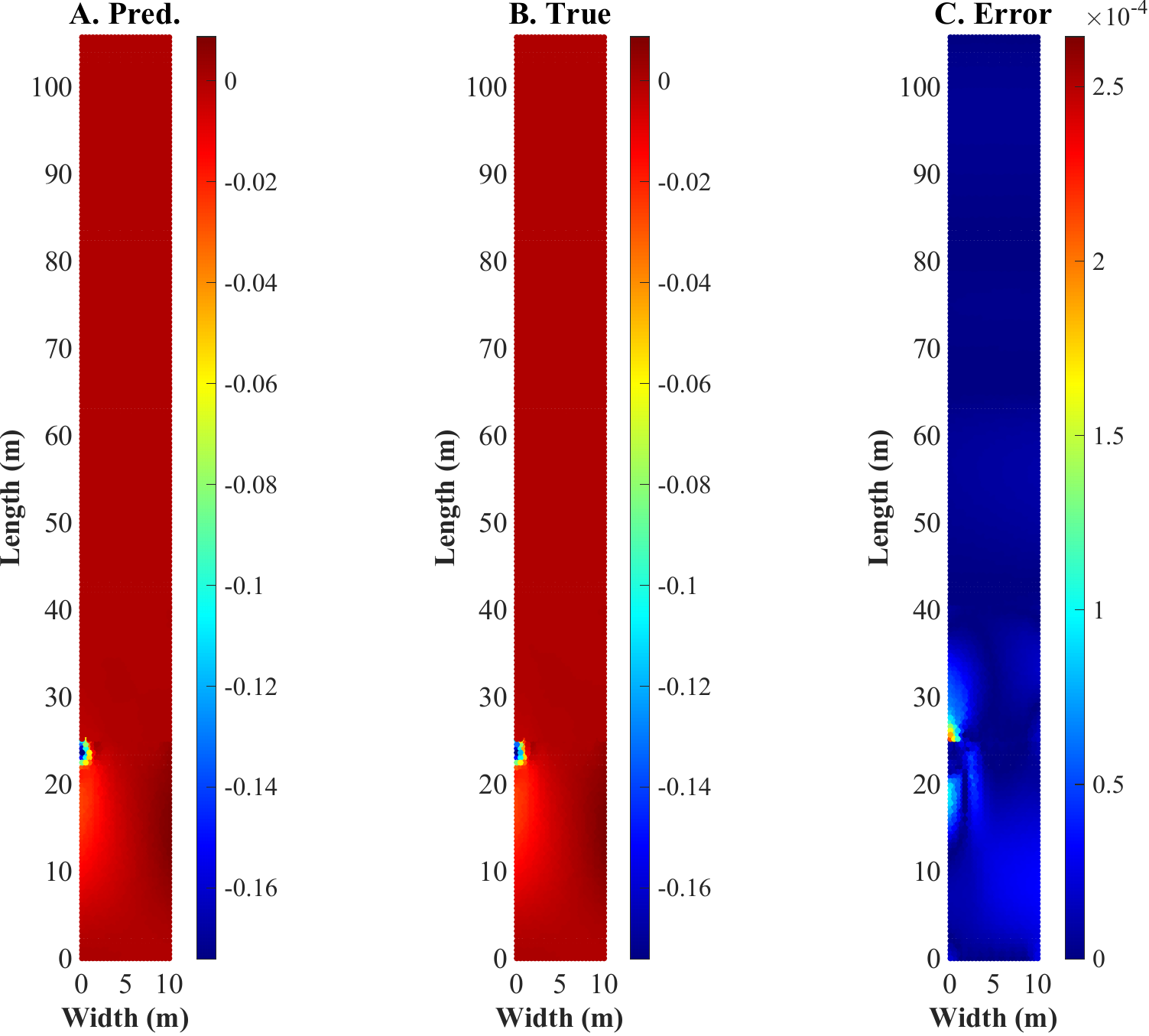}
    \caption{Single-load-case response comparison for the benchmark bridge deck in terms of vertical displacement ($U_y$). \textbf{A.} Predicted displacement field obtained from the proposed framework. \textbf{B.} Reference finite element solution. \textbf{C.} Absolute error distribution over the full structural domain.}
\label{Benchmark_Contours}
\end{figure}

\paragraph{Forward Analysis (Single Load Case)} 
For a given loading condition, the framework performs adaptive node selection, neural prediction, and physics-based reconstruction to recover the full-field structural response. Figure~\ref{Benchmark_Contours} presents a representative single-load case, comparing the predicted response, FEM reference solution, and absolute error contours for the transverse displacement $U_y$.

The results show strong agreement between the predicted and FEM responses, with error magnitudes typically 1--2 orders lower than the structural response values. The total time required to obtain a full-field prediction, including the stiffness-based reconstruction step, is approximately 0.3 seconds, compared to nearly 20 seconds for a single Abaqus simulation (Figure~\ref{Benchmark_Comparison} \textbf{B}). The operator inference (excluding post-processing) itself is up to four orders of magnitude faster, with the majority of the computational cost associated with the subsequent full-field reconstruction. This substantial reduction in computational effort enables rapid evaluation of multiple loading scenarios. Furthermore, the predicted single-wheel responses serve as the fundamental building block for reconstructing vehicle-level responses through the superposition framework described in Section~\ref{sec:wheel_decomposition}.

\paragraph{Influence Line and Surface Generation}

\begin{figure}[!htbp]
    \centering
    \includegraphics[width=0.9\textwidth]{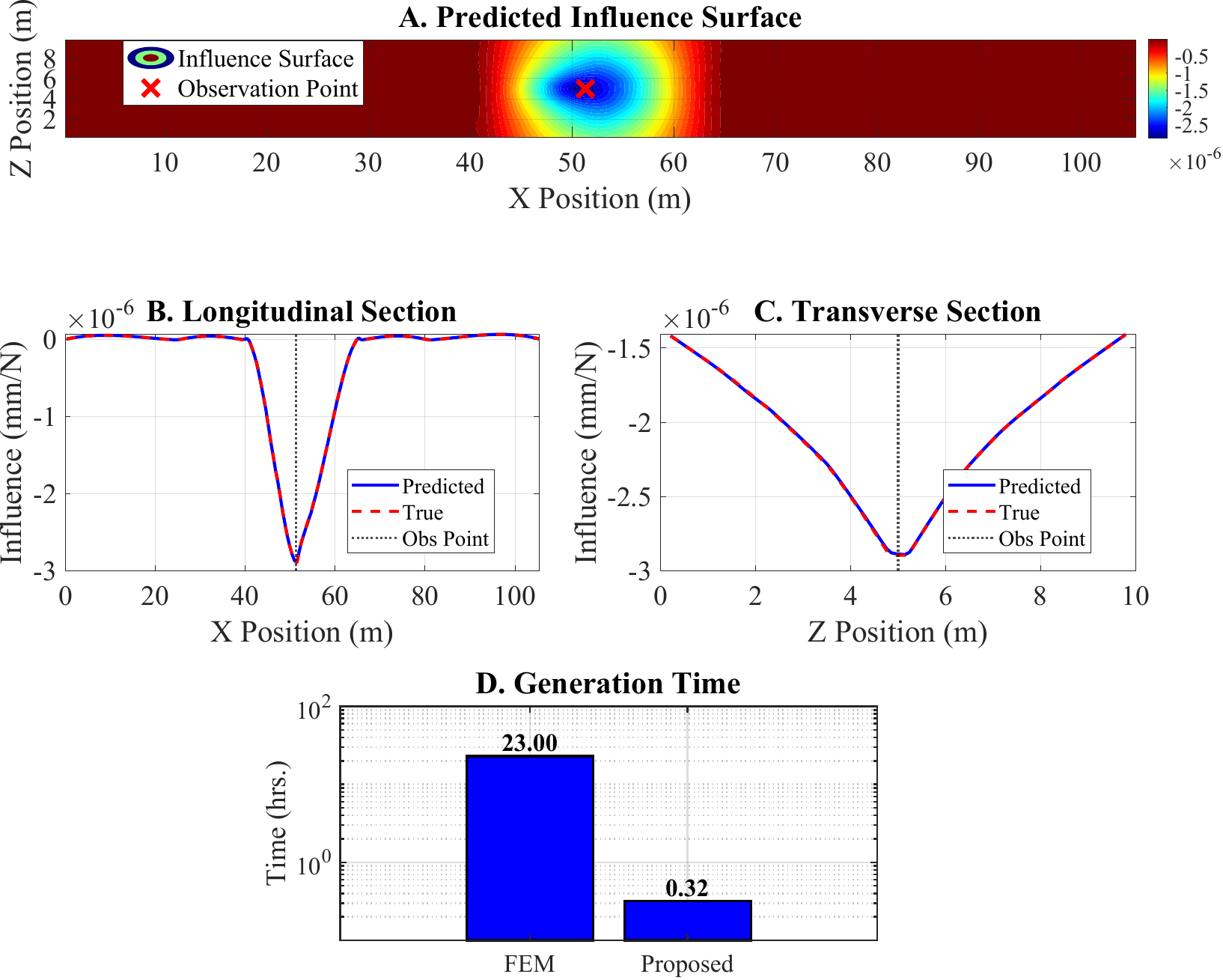}
    \caption{Influence response comparison in terms of vertical displacement ($U_y$). \textbf{A.} Predicted influence surface generated under a unit moving load at the selected observation location. \textbf{B.} Longitudinal influence line extracted at the observation section, compared against the reference FEM solution. \textbf{C.} Transverse influence line at the same observation location, together with the corresponding FEM comparison. \textbf{D.} Computational time comparison between the conventional FEM approach and the proposed framework for influence surface generation.}
\label{Benchmark_Influence}
\end{figure}

The proposed AD-DeepONet enables efficient generation of influence lines and influence surfaces through repeated operator evaluations. In the present study, approximately 3000 unit-load positions are evaluated by sweeping the load across the bridge deck, while the response at a selected observation point is extracted to construct the corresponding influence surface. Figure~\ref{Benchmark_Influence} presents the comparison between the predicted and FEM-based influence responses for the vertical displacement component ($U_y$), showing strong agreement across the entire bridge domain. The computational advantage becomes particularly significant for this large-scale repeated analysis task. Conventional FEM requires approximately 23 hours to generate the influence surface for a single observation point, whereas the proposed framework completes the same analysis in approximately 0.32 hours through rapid operator inference.

\begin{figure}[!htbp]
    \centering
    \includegraphics[width=0.92\textwidth]{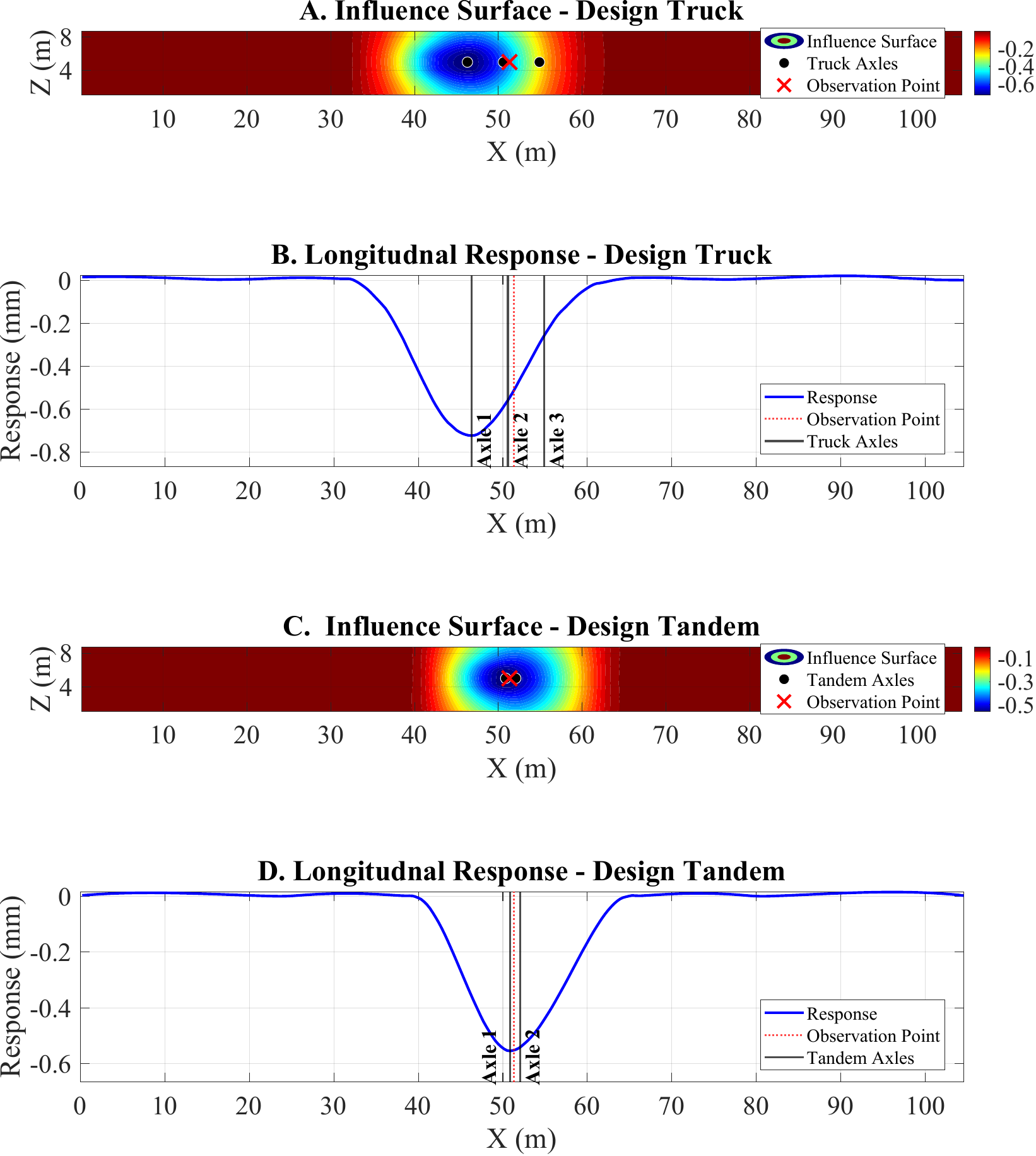}
    \caption{Influence response surfaces for design truck and tandem loading scenarios in terms of vertical displacement ($U_y$). \textbf{A.} Truck-load influence surface. \textbf{B.} Longitudinal response profile highlighting the critical axle locations corresponding to the maximum structural response. \textbf{C.} Tandem-load influence surface. \textbf{D.} Longitudinal response profile with the associated critical axle positions producing the peak response.}
\label{Benchmark_Truck}
\end{figure}

Design truck and design tandem configurations are subsequently constructed using standard axle layouts, where each vehicle is represented as a collection of wheel loads with prescribed spacing and magnitudes. The total structural response is then obtained through the superposition principle described in Section~\ref{sec:wheel_decomposition}. Figure~\ref{Benchmark_Truck} presents the resulting response distributions for the design truck and design tandem loading cases, together with the corresponding critical axle positions associated with the maximum response at the selected observation location.

\section{Mussafah Bridge}\label{sec:mussafah}

\subsection{Bridge Description}
\begin{figure}[!htbp]
    \centering
    \includegraphics[width=0.75\textwidth]{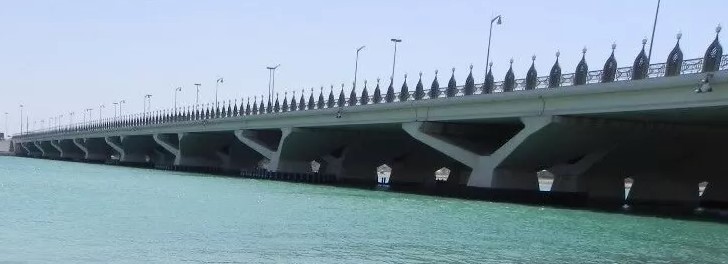}
    \caption{Mussafah Bridge, Abu Dhabi, UAE \cite{ivypanda_cnip}.}
    \label{Mussafah_Bridge}
\end{figure}

The Mussafah Bridge is considered as the primary large-scale target problem in this study and serves as the main motivation for the proposed AD-DeepONet framework. Located along Route E-20 in Abu Dhabi (Figure~\ref{Mussafah_Bridge}), the bridge is a prestressed concrete girder system consisting of multiple spans, expansion joints, heterogeneous structural components, and complex boundary conditions. The structure exhibits strong spatial variability in stiffness and load transfer mechanisms, resulting in highly localized responses that strongly depend on the vehicular load position. These characteristics make the Mussafah Bridge a challenging large-scale problem for operator learning under localized structural behavior.

\subsection{FEM Modeling and Data Generation}

\begin{figure}[!htbp]
    \centering
    \includegraphics[width=0.8\textwidth]{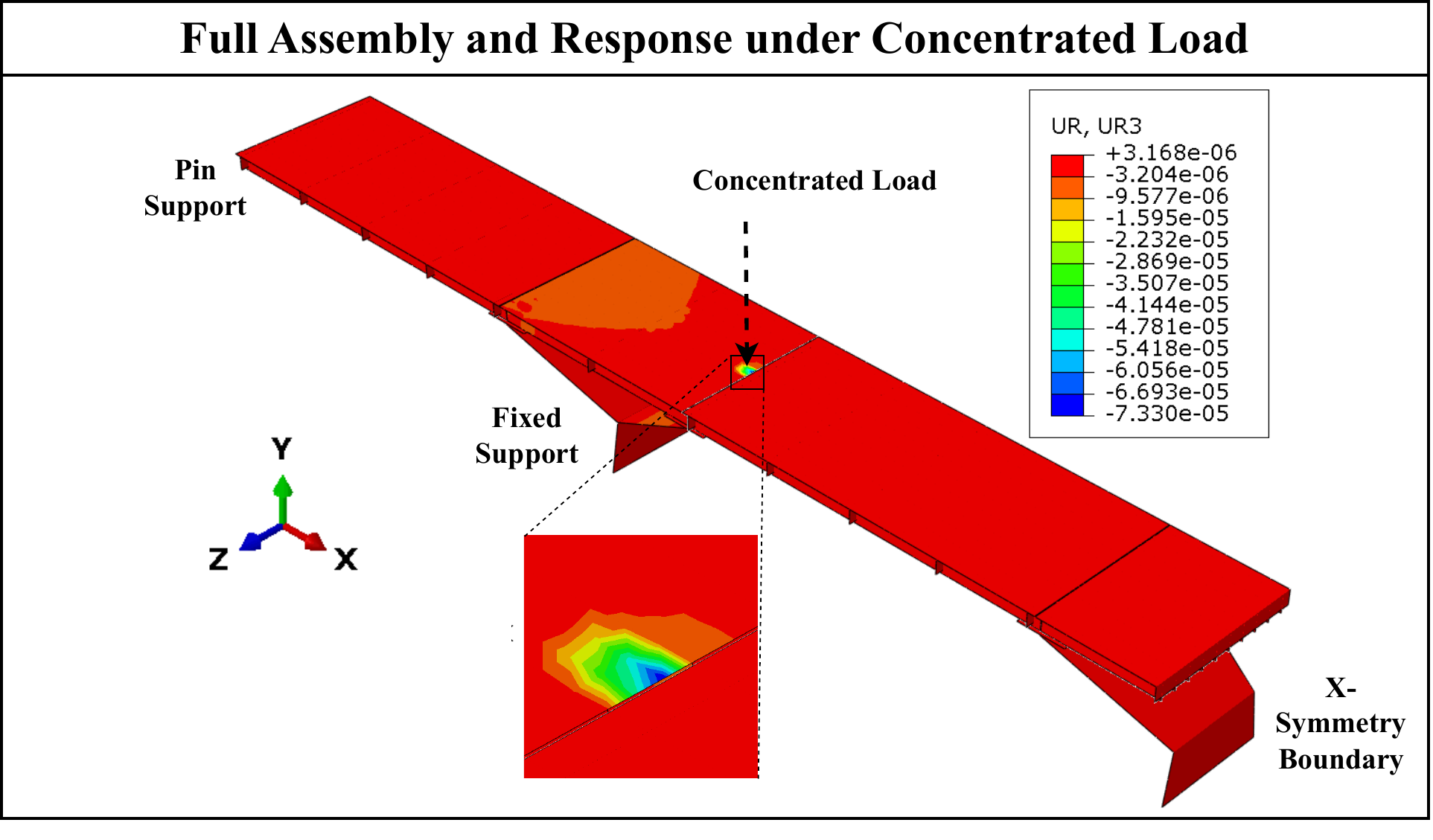}
    \caption{Reduced-order finite element representation of the Mussafah Bridge subjected to concentrated wheel loading, together with the corresponding structural response.}
    \label{Mussafah_simulation}
\end{figure}

To enable data-driven learning on this large-scale system, a finite element model is developed in Abaqus. A fully detailed 3D representation with solid elements and explicit reinforcement modeling is initially considered; however, such an approach becomes computationally intractable when extended to the full bridge due to the extremely large number of degrees of freedom. To address this, an equivalent shell reduced order modeling strategy is adopted (Figure \ref{fig:Shell}), where the heterogeneous structural components are represented using shell elements with homogenized material properties. These properties account for the combined effects of concrete, reinforcement, and prestressing, ensuring that the dominant global behavior is preserved while significantly reducing computational cost. The reduced-order representation is verified against detailed component-level models to ensure consistency in structural response. Further reduction is achieved by constructing the bridge model using representative segments rather than the full structure. This approach leverages structural repetition while retaining critical features such as span continuity, pier connections, and boundary transitions (Figure \ref{Mussafah_simulation}). The resulting model consists of approximately 15,050 nodes and 14,915 elements, corresponding to about 90,300 degrees of freedom, which remains sufficiently large to challenge standard learning methods.

A dataset of 40,000 loading scenarios is generated using the wheel-level loading framework introduced in Section~\ref{sec:wheel_decomposition}. Wheel loads are randomly distributed across the bridge deck, and the corresponding full-field structural responses are extracted for all translational and rotational degrees of freedom. Similar to the benchmark problem, the resulting responses remain highly localized in influence zone (Figure \ref{Mussafah_simulation}), producing a strongly imbalanced dataset dominated by near-zero responses across most of the structural domain. This makes the Mussafah Bridge an appropriate large-scale verification problem for evaluating localized operator learning.

\subsection{Methods Comparison}

\begin{figure}[!htbp]
    \centering
    \includegraphics[width=1\textwidth]{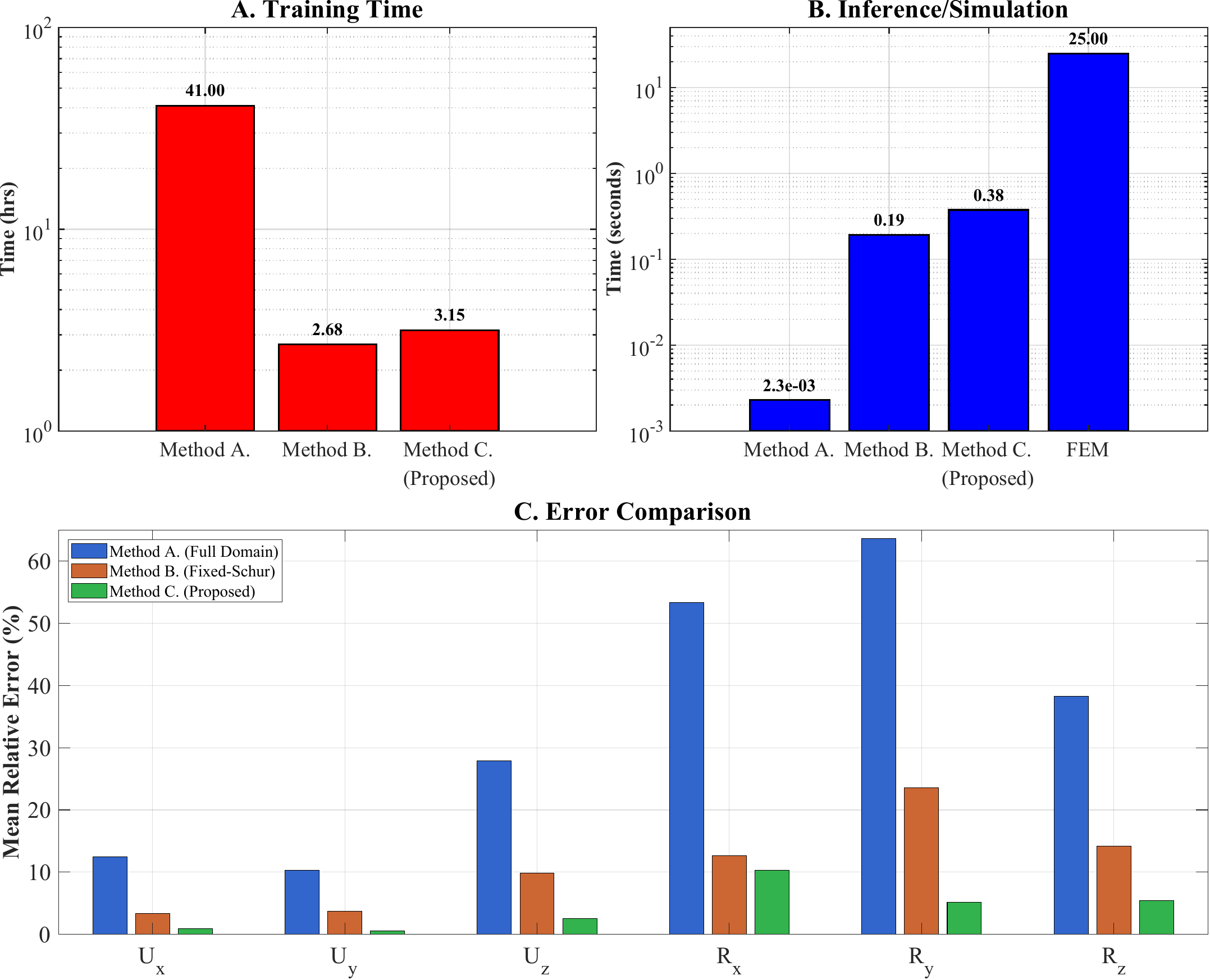}
    \caption{Comprehensive performance comparison of Methods A, B, and C for the Mussafah Bridge case study. \textbf{A.} Training time comparison. \textbf{B.} Inference time comparison. \textbf{C.} Mean relative error for each structural response component across all evaluated methods.}
    \label{Mussafah_Comparis}
\end{figure}

\begin{figure}[!htbp]
    \centering
    \includegraphics[width=1\textwidth]{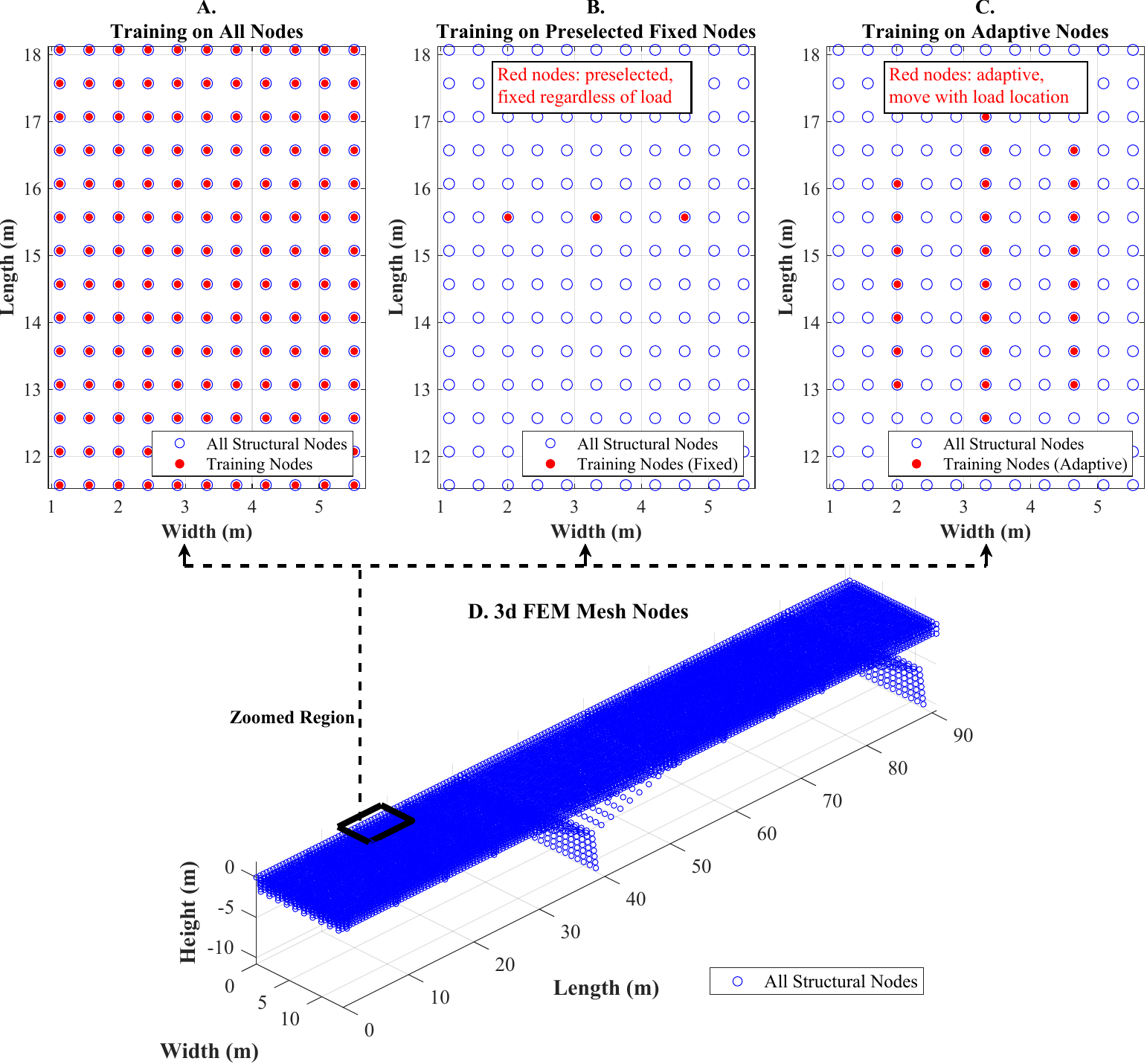}
    \caption{Comparison of training domains adopted for Mussafah Bridge. \textbf{A.} Method A (Vanilla DeepONet), where the entire structural domain is used for training and prediction. \textbf{B.} Method B (Fixed-Schur DeepONet \cite{ahmed2025physics}), which employs a predefined fixed Schur domain independent of the load location. \textbf{C.} Proposed Method C (Adaptive-Schur Domain), where the training domain dynamically adapts according to the load position, enabling the selected nodes to spatially align with the moving load. \textbf{D.} Three-dimensional finite element mesh representation of the Mussafah Bridge structure, highlighting the full computational domain.}
    \label{Mussafah_AllDomain}
\end{figure}

The three operator learning strategies are evaluated on the Mussafah Bridge to assess performance at the bridge scale. Figure \ref{Mussafah_Comparis} summarizes the training cost, inference time, and prediction errors, while Figure \ref{Mussafah_AllDomain} illustrates the corresponding learning domains. Similar trends to the benchmark study are observed, with Method C achieving the best overall balance between accuracy and computational efficiency.

\paragraph{Method A: Vanilla DeepONet (Full-Domain Learning)}

Method A produces the largest prediction errors and the highest computational cost. As shown in Figure \ref{Mussafah_All_Error} \textbf{A}, errors range from approximately 10\% to 65\% across different response components. The large output space also results in substantially longer training times compared to the reduced-domain approaches (Figure \ref{Mussafah_Comparis}).

\paragraph{Method B: Fixed Schur DeepONet (Ahmed et al.~\cite{ahmed2025physics})}

Method B significantly reduces computational cost while improving accuracy relative to Method A. However, noticeable errors remain, particularly for rotational components (Figure \ref{Mussafah_All_Error} \textbf{B}). Although the reconstructed full-field responses show lower global errors (Figure \ref{Mussafah_All_Error} \textbf{C}), the prediction quality remains less consistent.

\subsubsection{Method C: AD-DeepONet (Proposed)}

\begin{figure}[!htbp]
    \centering
    \includegraphics[width=0.95\textwidth]{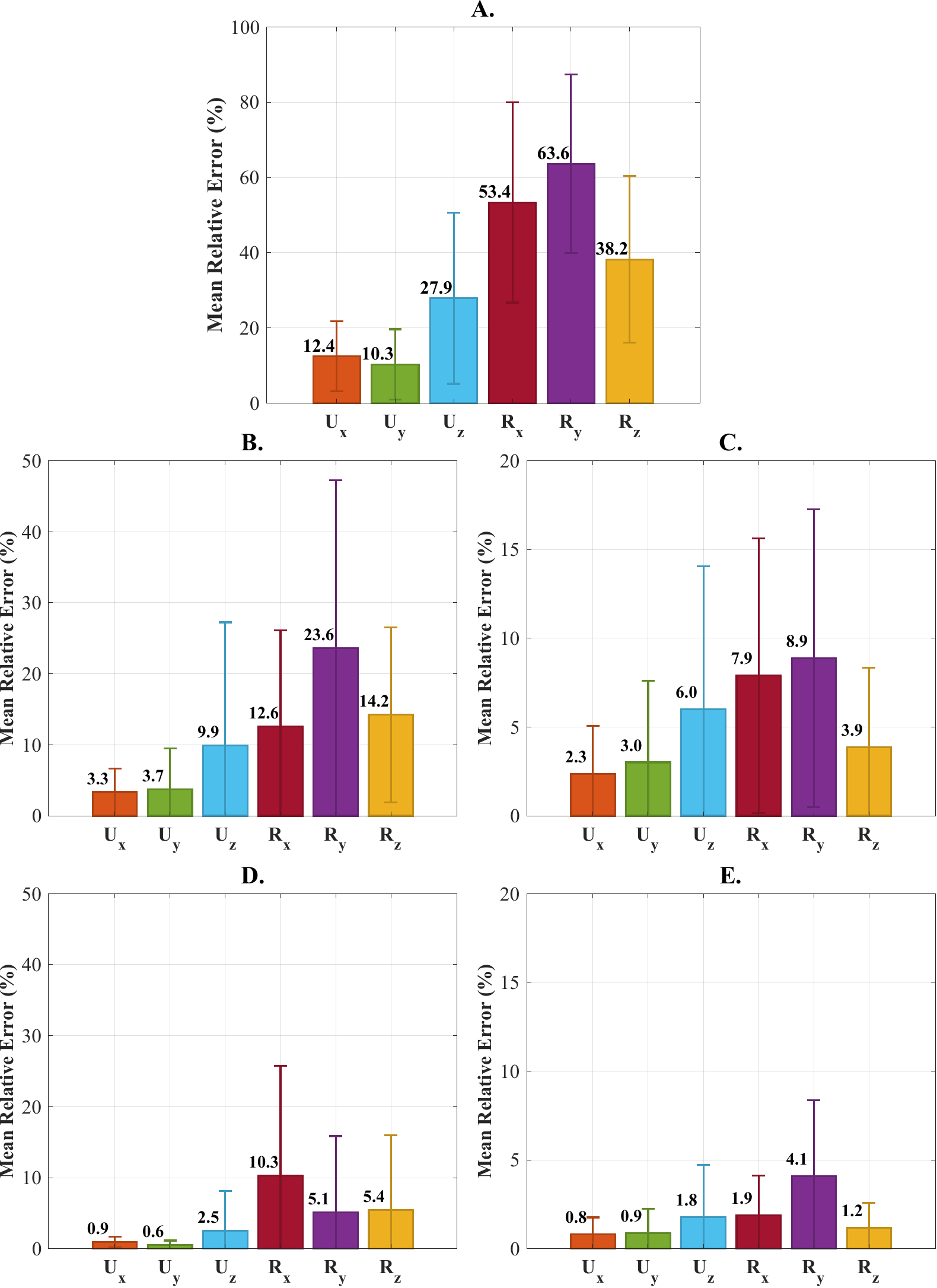}
    \caption{Prediction error comparison for Methods A, B, and C on the Mussafah Bridge case study. \textbf{A.} Full-domain prediction error for Method A (Vanilla DeepONet). \textbf{B.} Prediction error evaluated at the fixed Schur nodes for Method B (Fixed-Schur DeepONet). \textbf{C.} Corresponding full-field reconstruction error for Method B using the stiffness-based reconstruction framework. \textbf{D.} Prediction error evaluated at the adaptive Schur nodes for the proposed Method C (Adaptive-Schur DeepONet). \textbf{E.} Corresponding full-field reconstruction error for Method C using the stiffness-based reconstruction framework.}
    \label{Mussafah_All_Error}
\end{figure}

Method C achieves the best overall performance. As shown in Figure \ref{Mussafah_All_Error} \textbf{D}, prediction errors at adaptive nodes remain below 3\% for most displacement components. Following full-field reconstruction (Figure \ref{Mussafah_All_Error} \textbf{E}), the overall error remains below 5\%, with most displacement components below 2\%. These improvements are achieved while maintaining sub-second inference time and substantially lower training cost than Method A (Figure \ref{Mussafah_Comparis}), demonstrating the scalability of the proposed framework to real bridge-scale applications.

\subsection{Application}

\begin{figure}[!htbp]
    \centering
    \includegraphics[width=0.8\textwidth]{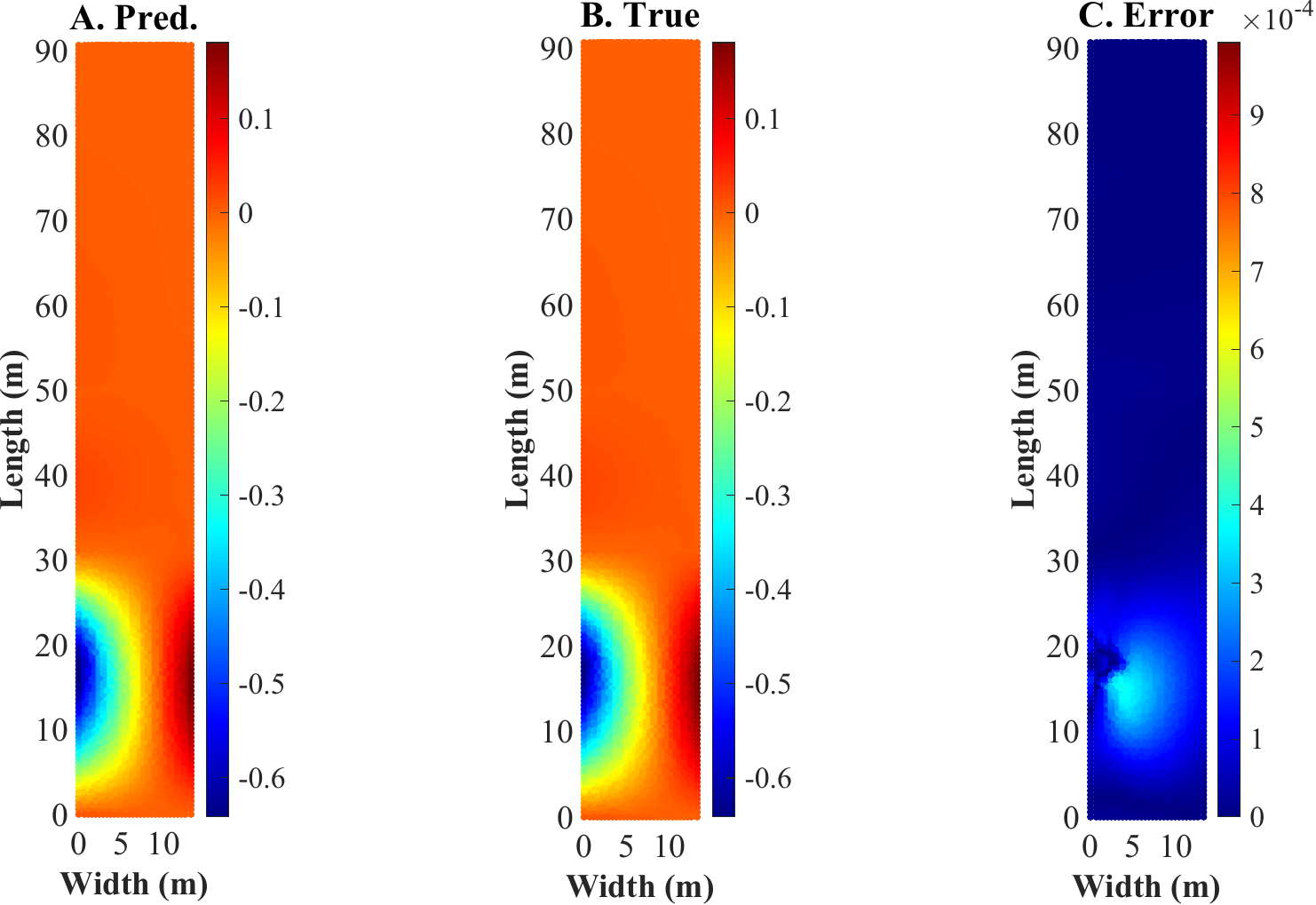}
    \caption{Single-load-case response comparison for the Mussafah Bridge deck in terms of vertical displacement ($U_y$). \textbf{A.} Predicted displacement field obtained from the proposed framework. \textbf{B.} Reference finite element solution. \textbf{C.} Absolute error distribution over the full structural domain.}
    \label{Mussafah_Contours}
\end{figure}

\paragraph{Single Load Case (Forward Analysis)} 
Figure~\ref{Mussafah_Contours} presents a representative single-load case, comparing the predicted response, FEM solution, and absolute error contours for the vertical displacement component ($U_y$). Excellent agreement is observed between the predicted and reference responses despite the increased structural complexity of the bridge. The total response evaluation time, including stiffness-based reconstruction, is approximately 0.4 seconds per load case compared to nearly 25 seconds for a conventional FEM simulation.

\paragraph{Influence Line and Surface Generation}

\begin{figure}[!htbp]
    \centering
    \includegraphics[width=1\textwidth]{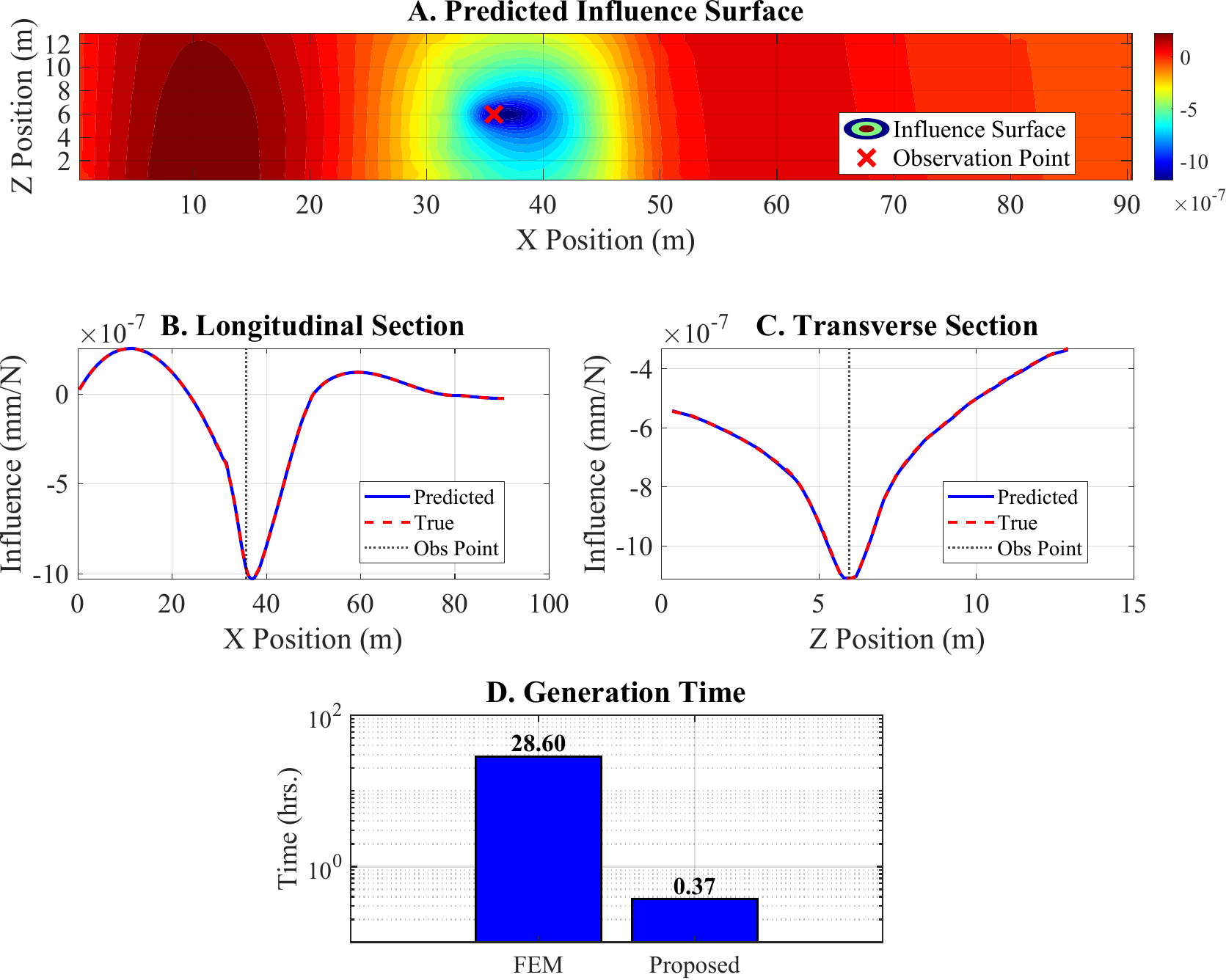}
    \caption{Influence response comparison in terms of vertical displacement ($U_y$) for Mussafah Bridge. \textbf{A.} Predicted influence surface generated under a unit moving load at the selected observation location. \textbf{B.} Longitudinal influence line extracted at the observation section, compared against the reference FEM solution. \textbf{C.} Transverse influence line at the same observation location, together with the corresponding FEM comparison. \textbf{D.} Computational time comparison between the conventional FEM approach and the proposed framework for influence surface generation.}
    \label{Mussafah_InfluenceComparison}
\end{figure}

The proposed framework is further applied to large-scale influence analysis by evaluating approximately 3300 unit-load positions across the bridge deck. Figure \ref{Mussafah_InfluenceComparison} compares the predicted and FEM-based influence responses, showing strong agreement over the entire loading domain. While conventional FEM requires approximately 28.6 hours to generate the influence surface for a single observation point, the proposed AD-DeepONet completes the same analysis in approximately 0.37 hours.

\begin{figure}[!htbp]
    \centering
    \includegraphics[width=0.92\textwidth]{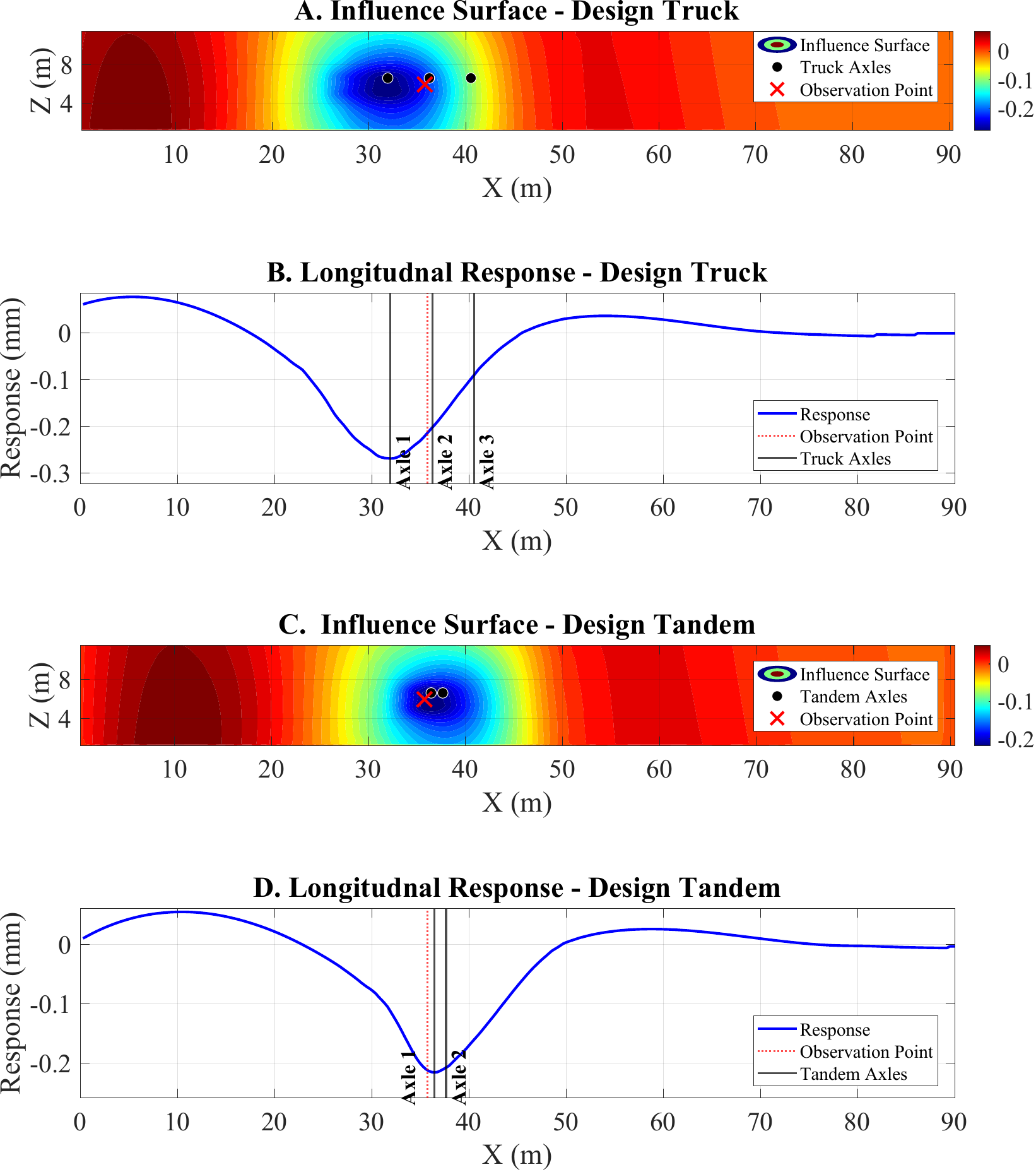}
    \caption{Influence response surfaces for design truck and tandem loading scenarios in terms of vertical displacement ($U_y$) for Mussafah Bridge. \textbf{A.} Truck-load influence surface. \textbf{B.} Longitudinal response profile highlighting the critical axle locations corresponding to the maximum structural response. \textbf{C.} Tandem-load influence surface. \textbf{D.} Longitudinal response profile with the associated critical axle positions producing the peak response.}
    \label{Mussafah_Truck}
\end{figure}

Figure \ref{Mussafah_Truck} presents the resulting influence responses for design truck and tandem load configurations together with the corresponding critical vehicle positions. These results demonstrate that the proposed framework can efficiently evaluate realistic multi-axle loading scenarios on large-scale bridge systems.

\section{Conclusions \& Limitations}\label{sec:conclusion}

This study presented a AD-DeepONet for efficient and scalable structural analysis of long-span roadway bridges, with the Mussafah Bridge serving as the primary target problem. The work addresses a fundamental limitation in operator learning for structural systems, namely the inability of conventional approaches to capture highly localized responses within large spatial domains. The proposed framework overcomes this challenge by introducing a load-dependent learning domain, where the model dynamically focuses on the influence zone of the dominant structural activity. Combined with distance-aware feature representation and stiffness-based reconstruction, this enables accurate prediction of full-field structural response while maintaining physical consistency. The results demonstrate that conventional full-domain learning (Method A) is not suitable for large-scale bridge systems with localized behavior, and that fixed reduced-domain approaches (Method B) remain insufficient due to their lack of adaptability. In contrast, the proposed adaptive formulation (Method C) achieves consistent accuracy across the domain, with errors typically within 5\% relative to FEM, while significantly reducing computational cost. In particular, the ability to perform rapid repeated evaluations enables efficient generation of influence lines and surfaces, which are otherwise computationally prohibitive using conventional methods. The proposed framework provides a practical pathway toward real-time structural analysis and digital twin applications for large bridge systems. By combining data-driven learning with physics-based reconstruction and adaptive domain representation, the method offers a scalable solution for problems where both accuracy and computational efficiency are critical. 

Despite these promising results, several limitations remain. The current study focuses on static linear structural response under concentrated loading. Extension to dynamic and time-dependent problems requires incorporation of temporal operator learning. In addition, the adaptive domain is currently constructed using geometric proximity through a KNN strategy and does not explicitly account for structural characteristics such as stiffness distribution, boundary conditions, or load paths. Future work will investigate physics-informed and graph-based domain selection strategies, transfer learning across different bridge typologies, and extension to nonlinear structural behavior including material degradation and damage evolution.
\setcounter{section}{0}
\renewcommand{\thesection}{A.\arabic{section}}

\setcounter{figure}{0} \renewcommand{\thefigure}{A.\arabic{figure}}

\section{Implementation Details of Adaptive DeepONet}\label{app:architecture}

\begin{figure}[!htbp]
    \centering
    \includegraphics[width=.95\textwidth]{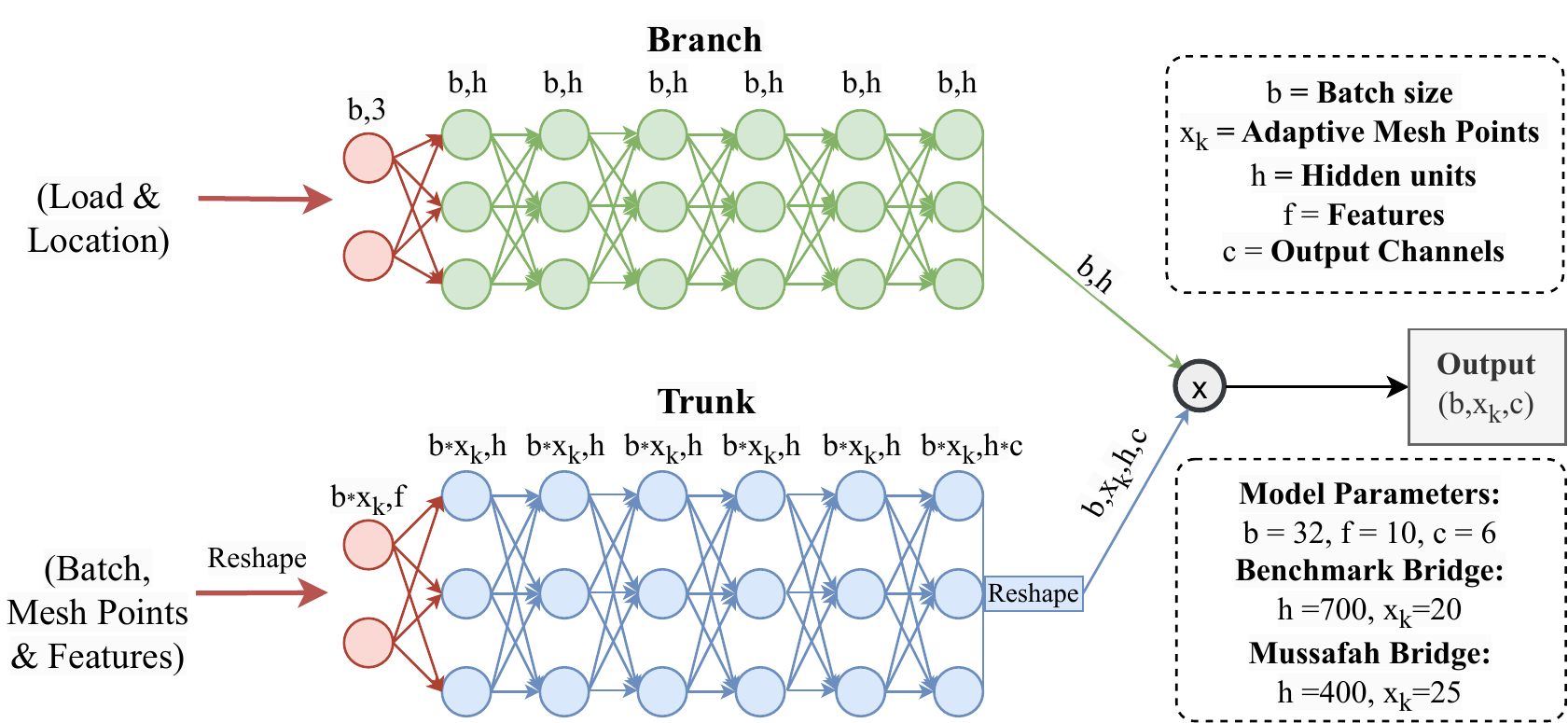}
    \caption{Architecture of the proposed adaptive DeepONet, illustrating branch and trunk network configurations along with corresponding data dimensionality}
    \label{Adaptive_Archi}
\end{figure}

Due to the sample-dependent adaptive Schur domain, the trunk input is constructed in a batched format of size $(batch \times x_k \times f)$, where $x_k$ denotes the number of selected nodes and $f$ represents the number of input features, including spatial coordinates and distance-aware features. For efficient processing using a standard multilayer perceptron (MLP), the trunk input is reshaped into a two-dimensional tensor of size $(batch \cdot x_k \times f)$. After passing through the trunk network, the output is reshaped back to $(batch \times x_k \times h \times c)$, where $h$ is the latent dimension and $c$ is the number of output channels corresponding to the structural degrees of freedom. The overall data flow through the network is illustrated in Figure ~\ref{Adaptive_Archi}.

The branch network encodes the loading condition, while the trunk network encodes the spatial features of the adaptive nodes. The final prediction is obtained through a tensor contraction between branch and trunk outputs, resulting in a tensor of size $(batch \times x_k \times c)$. This interaction is implemented using an Einsum formulation, enabling efficient computation across all adaptive nodes.

The adaptive formulation introduces additional preprocessing steps, including KNN-based node selection and feature construction for each sample. However, this overhead is negligible compared to full-domain learning. By reducing the effective output space, the approach significantly improves computational scalability while remaining compatible with stiffness-based reconstruction for full-field response recovery.

The network is trained using the Adam optimizer with a learning rate of $5 \times 10^{-4}$ and a batch size of 32. The dataset is split into 80\% for training and 20\% for testing, and this split is maintained consistently across all experiments. All computations are performed on the High-Performance Computing (HPC) cluster at NYU Abu Dhabi (Jubail campus), utilizing an Nvidia A100 GPU with 10 CPU cores.

\section{Effect of KNN Neighborhood Size}\label{sec:toy_parametric}

\begin{figure}[!htbp]
    \centering
    \includegraphics[width=0.8\textwidth]{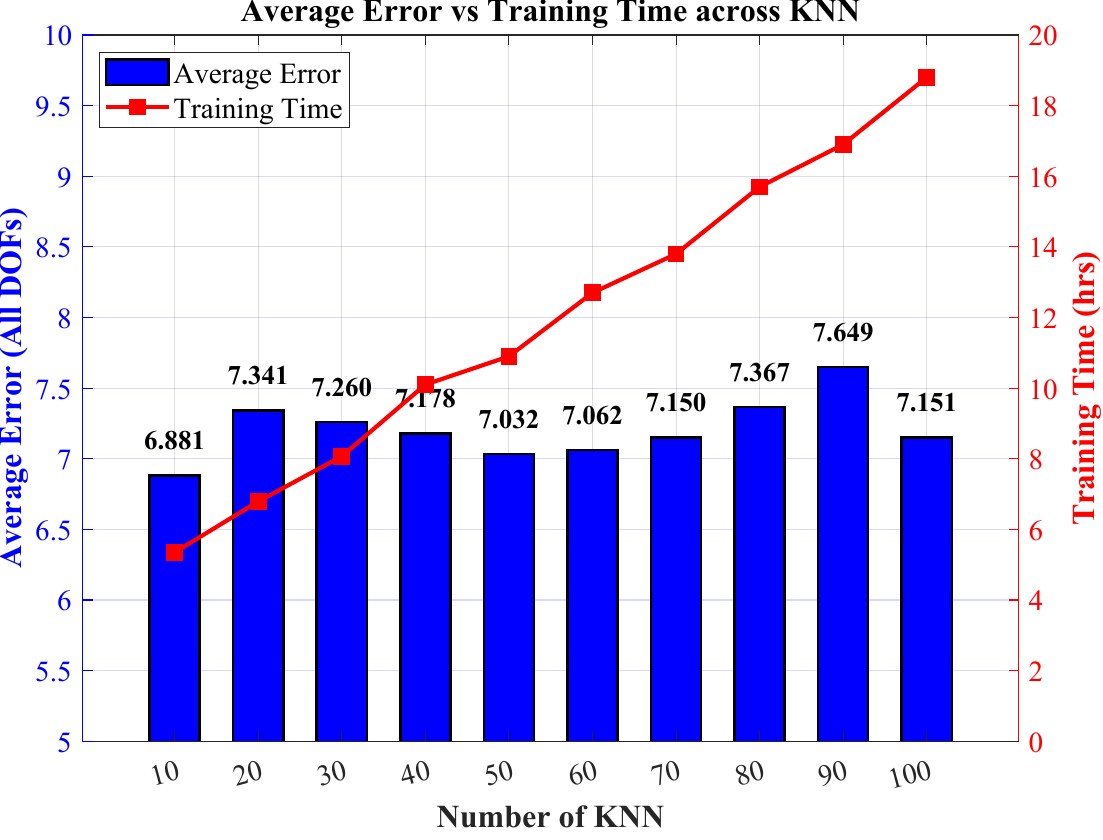}
    \caption{Effect of the number of KNN on prediction accuracy and training time for the adaptive Schur domain formulation in AD-DeepONet}
    \label{Benchmark_KNN}
\end{figure}

This study investigates the influence of the number of nearest neighbors ($K$) used in the adaptive Schur domain formulation described in Section \ref{sec:adpativesection}. In this formulation, each loading case defines a sample-dependent domain constructed using a KNN strategy, where $K$ controls the number of selected structural nodes. The choice of $K$ directly affects both the resolution of localized response capture and the computational efficiency of the model. Figure~\ref{Benchmark_KNN} presents the variation of prediction accuracy and training time with respect to $K$, ranging from 10 to 100 neighbors. The results show that the prediction accuracy remains largely stable across this range, with only minor variation in the mean relative error (approximately 6.9\% to 7.6\%). This indicates that the dominant localized structural response can be effectively captured even with a relatively small adaptive domain, as long as the selected nodes remain centered around the load location. In contrast, the computational cost is strongly affected by the choice of $K$. As the number of neighbors increases, the trunk network input grows proportionally, leading to a significant rise in training time. In particular, the training cost increases from a few hours at lower values of $K$ to nearly 19 hours when $K = 100$, primarily due to the expanded sample-dependent tensor operations and increased trunk dimensionality. Based on this trade-off, $K = 20$ is selected for the Benchmark bridge, as it provides a balanced compromise between accuracy, stability, and computational efficiency.

\section*{Credit authorship statement}
\textbf{Bilal Ahmed:} Conceptualization, Methodology, Software, Formal Analysis, Writing - Original draft, Data generation, Visualization, Supervision. \textbf{Diab Abueidda:} Conceptualization, Methodology, Software, Formal Analysis, Writing - Review and Editing. \textbf{Waleed El-Sekelly:} Supervision, Project administration.  \textbf{Tarek Abdoun:} Supervision, Project administration, Funding acquisition. \textbf{Mostafa Mobasher:} Conceptualization, Methodology, Writing - Review and Editing, Supervision, Project administration, Funding acquisition.

\section*{Acknowledgment}

This work was partially supported by the Department of Municipalities and Transport (DMT) under Grant No. DMT/CR/170/2025. This work was also partially supported by the Sand Hazards and Opportunities for Resilience, Energy, and Sustainability (SHORES) Center, funded by Tamkeen under the NYUAD Research Institute Award CG013. The authors wish to express their gratitude to the NYUAD Center for Research Computing for providing computational resources, services, and technical support.
\section*{Data availability}
All models and datasets associated with the benchmark study will be made publicly available upon acceptance of the manuscript.

 \bibliographystyle{elsarticle-num} 
 \bibliography{refs}

\end{document}